\documentclass[conference]{IEEEtran}
\usepackage{enumitem}
\usepackage{times}
\usepackage{listings}

\usepackage{amsmath,amsfonts,bm}

\def\eqref#1{equation~\ref{#1}}

\def\1{\bm{1}}

\DeclareMathAlphabet{\mathsfit}{\encodingdefault}{\sfdefault}{m}{sl}
\SetMathAlphabet{\mathsfit}{bold}{\encodingdefault}{\sfdefault}{bx}{n}

\usepackage[square,numbers,sort,compress]{natbib}
\usepackage[bookmarks=true]{hyperref}
\usepackage{url}
\usepackage{graphicx}
\usepackage{wrapfig}
\usepackage{duckuments}
\usepackage{booktabs}
\usepackage{tabularx}
\usepackage{multicol}
\usepackage{multirow}
\usepackage[dvipsnames]{xcolor}
\usepackage{capt-of}
\usepackage{caption}
\usepackage{cuted}

\usepackage{listings}

\DeclareFixedFont{\ttb}{T1}{txtt}{bx}{n}{12} %
\DeclareFixedFont{\ttm}{T1}{txtt}{m}{n}{12}  %

\usepackage{color}
\definecolor{deepblue}{rgb}{0,0,0.5}
\definecolor{deepred}{rgb}{0.6,0,0}
\definecolor{deepgreen}{rgb}{0,0.5,0}

\newcommand\pythonstyle{\lstset{
language=Python,
basicstyle=\ttm,
morekeywords={self},              %
keywordstyle=\ttb\color{deepblue},
emph={MyClass,__init__},          %
emphstyle=\ttb\color{deepred},    %
stringstyle=\color{deepgreen},
frame=tb,                         %
showstringspaces=false
}}

\lstnewenvironment{python}[1][]
{
\pythonstyle
\lstset{#1}
}
{}

\newcommand\pythoninline[1]{{\pythonstyle\lstinline!#1!}}

\title{CodeDiffuser: Attention-Enhanced Diffusion Policy via VLM-Generated Code for Instruction Ambiguity}

\author{\authorblockN{Guang Yin$^{3*}$\quad Yitong Li$^{4*}$\quad Yixuan Wang$^{1*}$\quad Dale McConachie$^{2}$\quad Paarth Shah$^{2}$\\Kunimatsu Hashimoto$^{2}$\quad Huan Zhang$^{3}$\quad Katherine Liu$^{2}$\quad Yunzhu Li$^1$}
\authorblockA{$^1$Columbia University\quad $^2$Toyota Research Institute \quad $^3$University of Illinois Urbana-Champaign \quad $^4$Tsinghua University}
\authorblockA{\textbf{\textcolor{magenta}{\url{https://robopil.github.io/code-diffuser/}}}}
}

\usepackage{listings}
\usepackage{color}

\definecolor{dkgreen}{rgb}{0,0.6,0}
\definecolor{gray}{rgb}{0.5,0.5,0.5}
\definecolor{mauve}{rgb}{0.58,0,0.82}

\begin{document}

\maketitle

\newcommand{\toyitong}{{\textcolor{red}{@ Yitong}}}
\newcommand{\toguang}{{\textcolor{orange}{@ Guang}}}
\newcommand{\toyixuan}{{\textcolor{blue}{@ Yixuan}}}
\newcommand{\yx}[1]{{\textcolor{blue}{[Yixuan: #1]}}}
\newcommand{\yxadd}[1]{{\textcolor{blue}{#1}}}
\newcommand{\huan}[1]{{\textcolor{brown}{[Huan: #1]}}}
\newcommand{\huanadd}[1]{{\textcolor{brown}{#1}}}
\newcommand{\katliu}[1]{{\textcolor{magenta}{[Katliu: #1]}}}
\newcommand{\katliuadd}[1]{{\textcolor{magenta}{#1}}}
\newcommand{\yz}[1]{{\textcolor{cyan}{[Yunzhu: #1]}}}
\newcommand{\yzadd}[1]{{\textcolor{cyan}{#1}}}
\newcommand{\guang}[1]{{\textcolor{orange}{[Guang: #1]}}}
\newcommand{\guangadd}[1]{{\textcolor{orange}{#1}}}

\makeatletter
\def\blfootnote{\gdef\@thefnmark{}\@footnotetext}
\makeatother

\setlist[itemize]{left=0mm}

\blfootnote{$^*$Denotes equal contribution.}
\begin{strip}
    \centering
    \includegraphics[width=\linewidth]{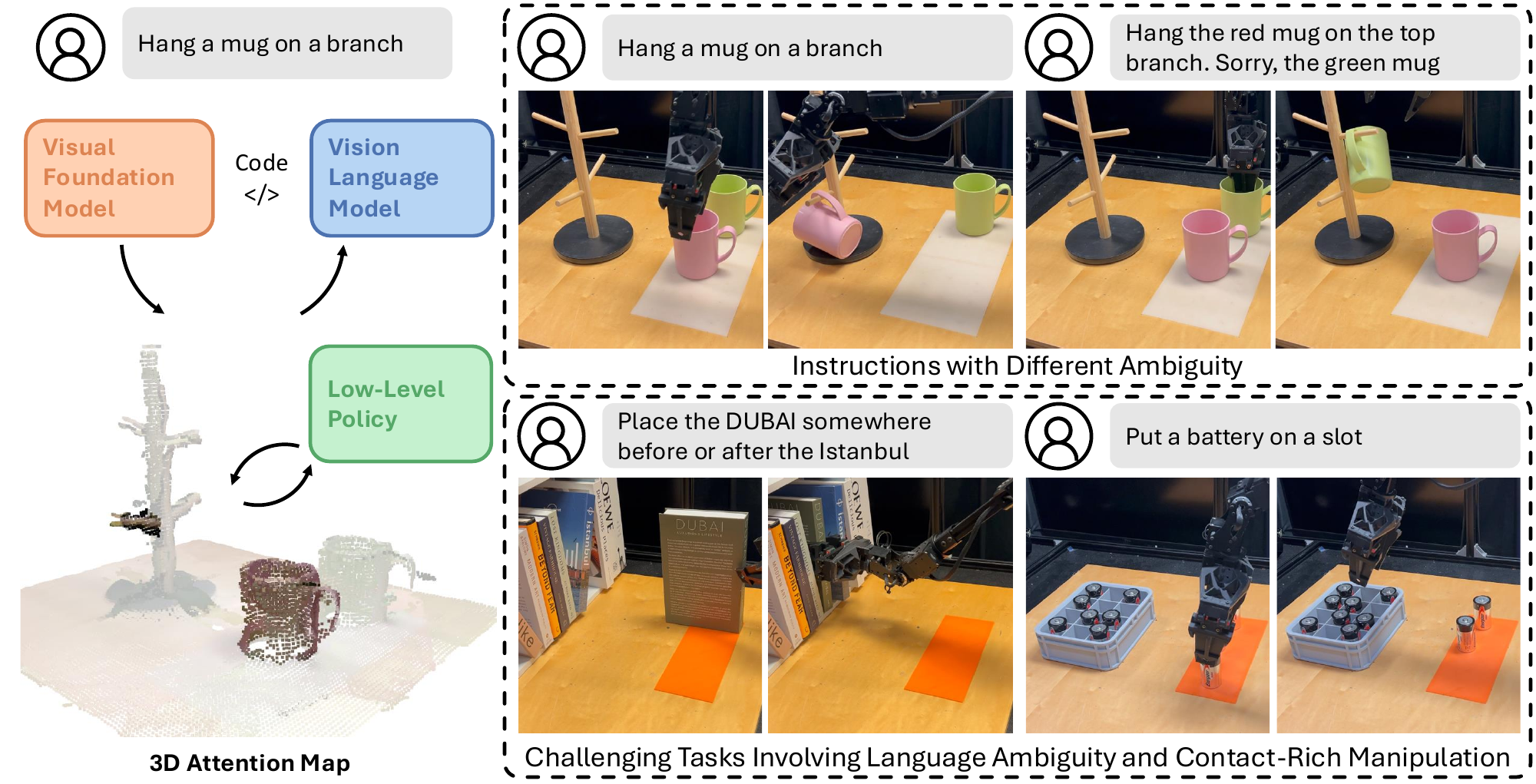}
    \vspace{-15pt}
    \captionof{figure}{\small
    \textbf{CodeDiffuser} leverages the code generated by Vision-Language Models (VLMs) as an interpretable and executable representation to understand abstract and ambiguous language instructions. This generated code interfaces with Visual Foundation Models (VFMs) to compute 3D attention maps, serving as an intermediate representation that highlights task-relevant areas and communicates with the low-level policy. Through extensive evaluations in both simulation and real-world settings, we demonstrate our method's effectiveness in challenging language-conditioned robotic tasks involving language ambiguity, contact-rich manipulation, and multi-object interactions.
    }
    \vspace{-5pt}
    \label{fig:teaser}
\end{strip}

\begin{abstract}

Natural language instructions for robotic manipulation tasks often exhibit ambiguity
and vagueness.
For instance, the instruction ``Hang a mug on the mug tree'' may involve multiple valid actions if there are several mugs and branches to choose from. Existing language-conditioned policies typically rely on end-to-end models that jointly handle high-level semantic understanding and low-level action generation, which can result in suboptimal performance due to their lack of modularity and interpretability. To address these challenges, we introduce a novel robotic manipulation framework that can accomplish tasks specified by potentially ambiguous natural language. This framework employs a Vision-Language Model (VLM) to interpret abstract concepts in natural language instructions and generates task-specific code — an interpretable and executable intermediate representation. The generated code interfaces with the perception module to produce 3D attention maps that highlight task-relevant regions by integrating spatial and semantic information, effectively resolving ambiguities in instructions. Through extensive experiments, we identify key limitations of current imitation learning methods, such as poor adaptation to language and environmental variations. We show that our approach excels across challenging manipulation tasks involving language ambiguity, contact-rich manipulation, and multi-object interactions.

\end{abstract}

\section{Introduction}

\begin{figure*}[htbp]
    \centering
    \includegraphics[width=\linewidth]{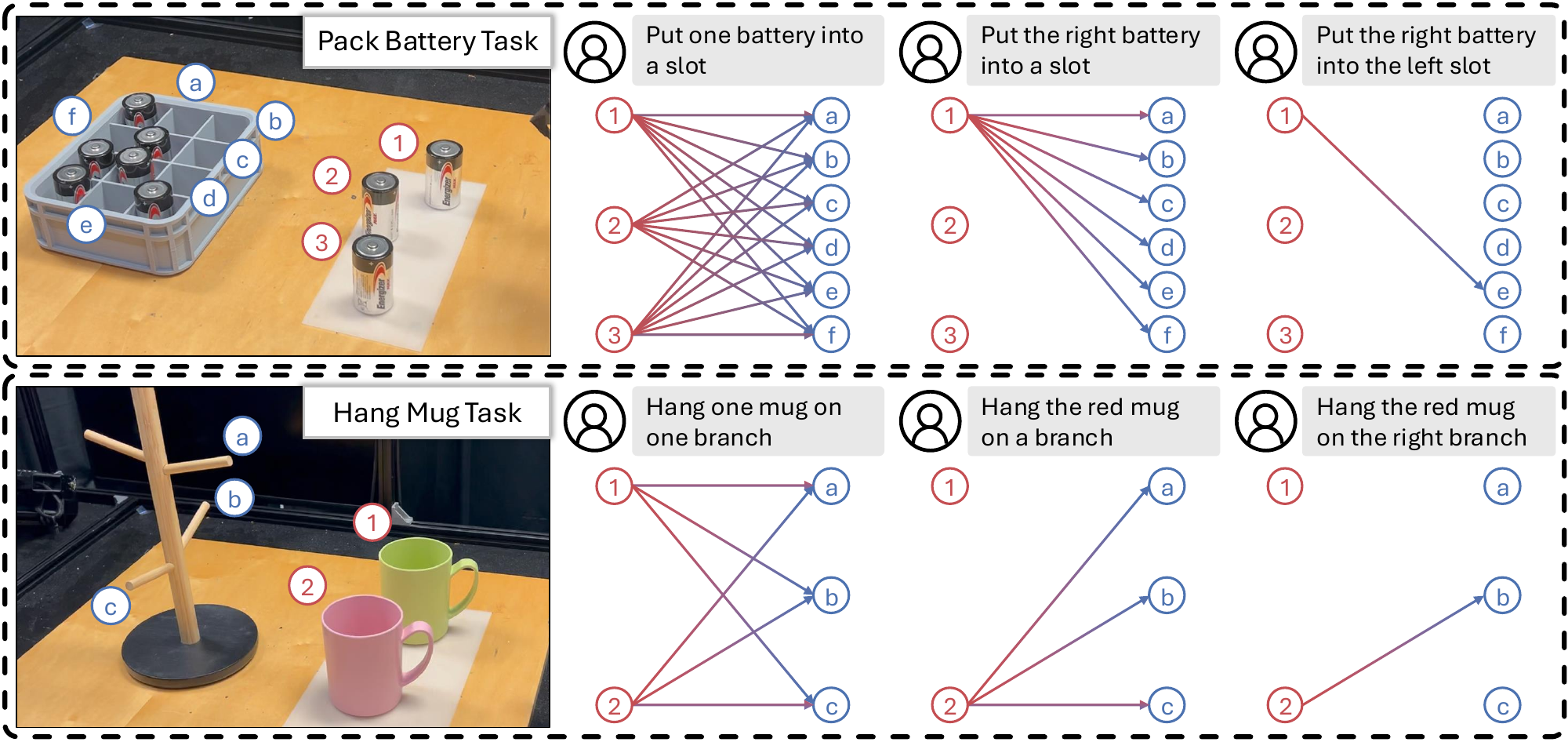}
    \vspace{-15pt}
    \caption{\small
    \textbf{Language Ambiguity for Task Specification.} Natural language instructions for robotic tasks often exhibit inherent ambiguity and vagueness. Consider the instruction ``Put one battery into a slot''--a common task in factory settings. In the given scenario, this instruction can be executed through multiple possible actions; the robot can choose from three available batteries and place one into any of the six potential slots in the tray, resulting in a total of eighteen possible choices. Furthermore, language instructions can vary in ambiguity, ranging from highly ambiguous directives to those that explicitly specify the target instance and goal location.
    }
    \label{fig:task_multi}
    \vspace{-10pt}
\end{figure*}

Natural language instructions are often vague and ambiguous when used to specify robotic tasks. As shown in Figure~\ref{fig:task_multi}, the instruction ``Pack the battery into the tray'' could involve multiple feasible execution paths. However, current language-conditioned imitation learning methods typically deploy end-to-end models to jointly handle high-level semantic understanding and low-level action prediction, which can lead to suboptimal performance.
In our experiments, we found that existing diffusion policies converge to a notably low success rate—well below practical usability—on challenging tasks involving language ambiguity, even with extensive amounts of data~\cite{chi2023diffusionpolicy, chi2024diffusionpolicy}.

In this work, we propose a novel robotic manipulation framework to accomplish tasks with potential language ambiguities.
Building upon the key technical insight that code generation provides an interpretable and executable interface between the visual-semantic understanding of Vision-Language Models (VLMs) and the dexterous physical capabilities of visuomotor policies, we introduce CodeDiffuser in this work to effectively handle language ambiguity and enhance semantic understanding capabilities.
Our system consists of three major components: code generation using VLMs, 3D attention computing using  Visual Foundation Models (VFMs), and low-level visuomotor policy.
The VLM first processes both visual observations and language descriptions and uses the provided perception API to generate code that produces a 3D attention map, highlighting task-relevant areas by leveraging visual features from foundational vision models such as DINOv2 and SAM~\cite{oquab2023dinov2, kirillov2023segany}. The 3D point cloud and attention map are then fed into the visuomotor policy, which generates an action trajectory to complete the task.

We conduct comprehensive experiments to analyze the limitations of state-of-the-art imitation learning algorithms when handling language ambiguities. Our results show that performance declines as language ambiguity increases. Furthermore, we demonstrate that simply increasing the number of demonstrations does not significantly enhance the ability to address language ambiguity and vagueness.

To gain deeper insights into attention map conditioning, we evaluate the language-conditioned 3D attention map and the visuomotor policy separately. Both quantitative and qualitative results show that 3D attention maps generated by VLM-based code accurately align with human instructions and effectively highlight task-relevant locations.
Furthermore, extensive quantitative analysis demonstrates that the 3D attention map serves as an effective representation, improving the visuomotor policy’s ability to handle language ambiguity. Finally, we evaluate the entire system and show that CodeDiffuser successfully performs challenging robotic manipulation tasks
with language ambiguity and vagueness.

In summary, our contributions are threefold:
1) We systematically evaluate the key limitations of state-of-the-art imitation learning frameworks in scenarios with language ambiguity and vagueness, such as pick-and-place tasks with multiple possible objects and destinations.
2) We propose CodeDiffuser, a novel robotic manipulation framework that addresses these challenges using VLM-generated code as an interpretable and executable intermediate representation. By interfacing with perception APIs, it generates 3D attention maps to bridge visual-semantic reasoning and low-level trajectory prediction.
3) We conduct extensive evaluations of individual modules and the full system in both simulation and real-world tasks, including contact-rich 6-DoF manipulation with multi-object interactions, demonstrating the effectiveness of our approach in handling language ambiguity.

\section{Related Works}

\begin{figure*}[htbp]
    \centering
    \includegraphics[width=\linewidth]{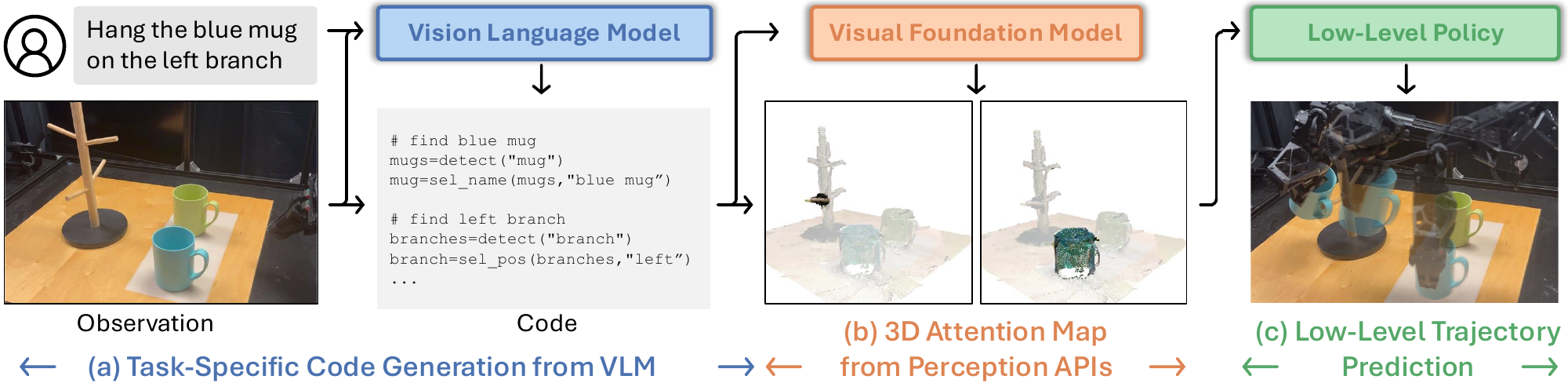}
    \vspace{-10pt}
    \caption{\small
    \textbf{Method Overview.} CodeDiffuser consists of three primary components: code generation, 3D attention map computation, and low-level policy. (a) CodeDiffuser first leverages the VLM semantic reasoning and code generation capabilities to understand human instructions with potentially ambiguity and environmental observations, generating task-specific code. (b) This generated code interfaces with perception APIs, built upon the VFM, to compute 3D attention maps that highlight task-relevant areas. (c) The attention maps are then fed into a low-level policy to generate actions accomplishing tasks with multi-object interaction, contact-rich manipulation, and language ambiguity.
    }
    \label{fig:method}
    \vspace{-10pt}
\end{figure*}

\subsection{Imitation Learning for Robotic Manipulation}

Imitation learning has achieved impressive results in dexterous robotic manipulation tasks, such as spreading sauces, flipping mugs~\cite{chi2023diffusionpolicy, chi2024diffusionpolicy}, tying shoelaces~\cite{zhaoaloha}, and more~\cite{zhao2023learning, fu2024mobile, chi2024universal, ha2024umi, zhang2018deep, goyal2023rvt, shridhar2022peract, shridhar2021cliport, zhu2022viola, robomimic2021, florence2019self, ross2011reduction, duan2017one, shi2023waypoint, di2024keypoint, di2024dinobot, wen2022you, padalkar2023open, brohan2022rt, brohan2023rt, ganapathi2022implicit, florence2021implicit, shafiullah2022behavior, du2019implicit, Welling2011BayesianLV, mordatch2018concept, octo_2023, pearce2023imitating, Ze2024DP3, kerr2024robotrobotdoimitating}.
While existing imitation learning frameworks model policies in an end-to-end manner—requiring them to jointly understand high-level semantics and predict low-level skills—this often results in suboptimal performance in complex scenarios where instructions lack specificity.
In contrast, our framework leverages the visual-semantic reasoning capabilities of VLMs to interpret potentially vague or abstract natural language instructions and generate code—an interpretable and executable intermediate representation—that interfaces with perception APIs to provide structured inputs to the low-level policy.

\subsection{Foundational Vision Model for Robotics}

Foundational vision models have demonstrated impressive zero-shot generalization capabilities in 2D vision tasks such as detection, segmentation, and visual representation~\cite{oquab2023dinov2, radford2021learning, liu2023grounding, cheng2024yolo}. Many robotic systems have adopted these models due to their strong generalization ability~\cite{wangd, wanggendp, huangrekep, huang2023voxposer, qiu-hu-2024-geff, ze2023gnfactor, ke20243d, zhu2023learning, di2024keypoint, di2024dinobot, lin2024spawnnet, kerr2024robotrobotdoimitating, gu2024conceptgraphs, jatavallabhula2023conceptfusion, jiang2024roboexp, hughes2022hydra, maggio2024clio, hughes2024foundations, strader2024indoor, peng2023openscene, ha2022semantic, yokoyama2024vlfm, dai2024think}. Among these, GenDP and 3D Diffuser Actor are most relevant to our work~\cite{wanggendp, ke20243d}.
While GenDP demonstrates category-level generalization, it lacks the ability to understand natural language instructions. In contrast, our framework is capable of understanding potentially ambiguous natural language instructions by using visual-semantic reasoning capabilities of VLM and generated code as an intermediate representation.

The 3D Diffuser Actor introduces a diffusion model that predicts end-effector keyposes from point clouds augmented with visual features from foundational vision models.
In contrast, our approach uses VLM-generated code to compute 3D attention map, which highlights task-relevant regions and possesses much lower dimension compared to 3D Diffuser Actor for easier visuo-motor policy training. %
Furthermore, 3D Diffuser Actor predicts end-effector keyposes rather than a continuous trajectory, which requires manual keypose definition for each task and is insufficient for tasks requiring smooth trajectory control, such as stowing books. In contrast, our low-level visuomotor policy predicts a smooth trajectory, providing greater flexibility for diverse tasks.

\subsection{Code Generation for Robotics}

Due to their advanced visual-semantic reasoning and code generation capabilities, LLMs and VLMs have been widely applied to generate code for various robotic tasks, including manipulation~\cite{huang2023voxposer, driess2023palm, ahn2022can, huang2022inner, liang2023code, huangrekep, huang2024copa, liu2024moka, nasiriany2024pivot, du2023video, hong20233d, chen2024spatialvlm, gao2024physically, wang2024grounding, hsu2023ns3d, yuan2024robopoint, duan2024manipulate}, navigation~\cite{dai2024think, chen20232}, and planning~\cite{yang2024llm, hu2023look}, as reviewed in several surveys~\cite{hu2023toward, firoozi2023foundation, kawaharazuka2024real, yang2023foundation}. Existing works typically employ motion planning to generate trajectories~\cite{huangrekep, huang2023voxposer}. While these methods demonstrate impressive zero-shot performance, they lack the ability to learn skills from human demonstrations.
In contrast, our framework connects a VLM to the low-level visuomotor policy via VLM-generated code, seamlessly combining the high-level semantic reasoning capabilities of VLM with the smooth low-level control enabled by learned policies.

Among LLM- and VLM-based approaches applied to robotics, VoxPoser and ReKep~\cite{huang2023voxposer, huangrekep} are closely related to our work. They use a 3D voxel heatmap and 3D relational keypoints, respectively, as intermediate representations to encode both geometric and semantic information, incorporating various optimization methods to generate action trajectories for downstream manipulation tasks. However, for certain tasks—such as stowing books, which involve multi-object interactions—instantiating an optimization problem can be challenging due to the complex dynamics involved. In contrast, our approach learns a visuomotor policy from demonstrations, offering greater flexibility in handling a wide range of tasks.

\section{Method}

\subsection{Problem Statement}
\label{sec:prob}

Our learning goal is to match the distribution given in a dataset of collected demonstrations, i.e.,
\begin{align}
    \min_\pi L(p_{\text{pred}}(a|o_t, l), p_{\text{gt}}(a|o_t, l)),
\end{align}
with observations $o \in \mathcal{O}$, task descriptions $l \in \mathcal{L}$, and actions $a \in \mathcal{A}$. 
$p_{\text{pred}}(a|o_t, l)$ represents the predicted action distribution and $p_{\text{gt}}(a|o_t, l)$ denotes the ground truth action distribution provided by a human demonstration dataset. The action distribution $p(a|o_t, l)$ can be decomposed as follows:
\begin{align}
    p(a|o_t, l) &= \int_{z_t} p(a|o_t, l, z=z_t)p(z=z_t|o_t, l) \\
    &= \int_{z_t} p(a|z=z_t)p(z=z_t|o_t, l),
\end{align}
where $z_t$ is a task-relevant latent representation of the state such that $p(a|o_t, l, z=z_t)=p(a|z=z_t)$, i.e., $z_t$ contains enough information about the observation and instruction to predict the action.

We observe that in the presence of language ambiguity, the distribution $p(a|o_t, l)$ is highly multi-modal. While single-task imitation learning policies have recently shown great success in modeling the multi-modal action distributions~\cite{chi2024diffusionpolicy}, ambiguity in the instruction introduces an additional complexity. For instance, in the packing battery task illustrated in Figure~\ref{fig:task_multi}, if the instruction is ``Hang one mug on one branch'' without specifying the mug or branch instance, the probability of each battery-slot pair is $1/18$, imposing an additional axis of multi-modality in the language instructions. The highly multimodal distribution poses challenges for training an end-to-end policy to model $p(a|o_t, l)$. In practice, we find that learning a single end-to-end policy is suboptimal when dealing with task ambiguity. Notably, we show in Section~\ref{sec:exp_1} that the current state-of-the-art methods can fail to achieve a high success rate even with extensive training demonstrations.

Under this framework, we model $p(a|z=z_t)$ with modern generative imitation learning methods and $p(z=z_t|o_t,l)$ with a VLM. We adopt a latent representation indicating task-relevant areas to accomplish the task, i.e., a 3D attention map highlighting a specific mug-branch pair. We show that this representation can be used to condition the imitation learning algorithms, and can also be generated from language by leveraging the reasoning capabilities of VLMs. We first generate intermediate code from the instruction $l$ and multi-view RGBD observations $o_t \in \mathbb{R}^{K \times H \times W \times 4}$, where $K$ is the number of camera views, and $H$ and $W$ represent the image height and width, respectively, as described in Section~\ref{sec:code}. In Section~\ref{sec:attn}, we describe the API provided to the code generation process used to construct our state representation $z_t$, a 3D attention map that highlights task-relevant regions. Finally, this 3D attention map is input into the low-level policy, which predicts a sequence of 6D end-effector poses $a$ for the robot, as illustrated in Section~\ref{sec:policy}.

\subsection{Code Generation}
\label{sec:code}

CodeDiffuser uses a VLM to translate a natural language instruction into executable code that generates 3D attention maps for downstream consumption by the visuomotor policy.
In our work, we leverage the VLM's few-shot generalization and commonsense reasoning capabilities to generate perception code from a few provided examples.
We first provide a VLM, such as ChatGPT-4o~\cite{achiam2023gpt}, with several in-context example code snippets
for constructing 3D attention maps using our perception APIs. At inference time, we input a task description into the VLM, which then produces the code to create the 3D attention map. The generated code takes as input the RGB observations and outputs a 3D attention map. Importantly, the code generation process can generate code to further invoke VLMs, enabling more advanced semantic reasoning.
For instance, as shown in Figure~\ref{fig:method}~(a), if the instruction is ``Hang the blue mug on the left branch,'' the generated code first detects all instances of mugs. It then selects the ``blue mug'' instance from the detected mugs using another call to the VLM. A similar process is applied for branches, with the key difference being that instance selection for branches relies more on spatial relations, such as ``left'' rather than semantic attributes like ``blue''. The implementation of perception APIs is detailed in Section~\ref{sec:attn}.

\subsection{API for 3D Attention Map Generation}
\label{sec:attn}
To provide the VLM with an API to bridge the semantic meaning of an instruction and the corresponding 3D observation, we design a simple yet effective API, including the functions \texttt{detect}, \texttt{sel\_name}, and \texttt{sel\_pos}. Example usage of these functions is provided in-context during inference.

\textbf{\texttt{detect}}: The \texttt{detect} function takes multi-view RGBD observations from cameras with known intrinsics and extrinsics and an object name as input and outputs a list of object instances belonging to the object category. The detection pipeline includes 3D feature extraction and object clustering. We first extract 2D feature maps $\mathcal{W}_i \in \mathbb{R}^{H \times W \times F}$ of each view using the DINOv2 model~\cite{oquab2023dinov2}, where $F$ is the semantic feature dimension. Utilizing the depth images and camera parameters from all $K$ viewpoints, we fuse the 2D DINOv2 features into 3D space to obtain 3D point clouds $\mathcal{P}\in \mathbb{R}^{M\times 3}$ and associated 3D features $\mathcal{F}\in \mathbb{R}^{M\times F}$, where $M$ is the number of points.
In practice, we follow the implementation introduced in D$^3$Fields~\cite{wangd} to obtain the 3D point clouds and their associated semantic features.

From a reference object, we annotate a set of relevant images, obtaining a reference DINOv2 feature denoted as $\mathcal{F}_{\text{ref}} \in \mathbb{R}^F$. We observe that the annotation process is lightweight and is only required when first setting up a task for training. By comparing reference feature $\mathcal{F}_{\text{ref}}$ with 3D semantic features $\mathcal{F}$, we can compute the similarity $\mathcal{S} \in \mathbb{R}^M$, where each element in $\mathcal{S}$ represents how likely the point belongs to the object category. Given the similarity $\mathcal{S}$, we can cluster the point clouds from the same object category into different object instances using the density-based spatial clustering of applications with noise (DBSCAN). Each clustering centroid represents one object instance.

\textbf{\texttt{sel\_pos}}: 
This function selects the target instance from an input object list (obtained via \texttt{detect}) based on spatial relations, such as \textit{far away}, \textit{close to}, or \textit{left}. It is necessary for instructions requiring geometric understanding.
To implement this function, we invoke a VLM, which generates code to compute distances or compare coordinates to identify the target instance.

\begin{figure*}
    \centering
    \includegraphics[width=\linewidth]{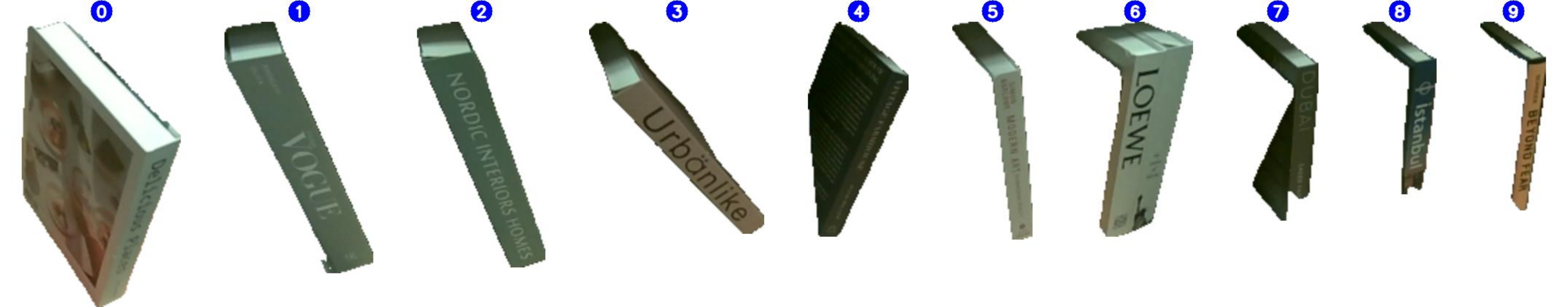}
    \caption{\small
    \textbf{Image Input to \texttt{sel\_name}.} After segmenting 3D instances and projecting them into 2D images, we concatenate the masked images and overlay instance labels. The \texttt{sel\_name} function takes this composite image as input and outputs the selected instance ID. This example shows the input image for the book stowing task.
    }
    \label{fig:inst_sel}
    \vspace{-15pt}
\end{figure*}

\textbf{\texttt{sel\_name}}: Similar to \texttt{sel\_pos}, but for instructions requiring semantic understanding, such as \textit{Bob's mug} or \textit{blue mug}. Inside this function, it passes the current observations into a VLM for instance selection. Concretely, we project 3D instances to 2D images and plot their corresponding bounding boxes. Then we feed the current observation overlayed with instance bounding boxes into VLM for instance selection.

Given the selected object instances, we project them onto 2D images and prompt Segment Anything~\cite{kirillov2023segany} using the projected 2D bounding boxes. Next, we fuse the 2D segmentations from multiple views into 3D space to generate a 3D attention map $\mathcal{I} \in \mathbb{R}^{M}$, where $M$ is the number of points. Each element in $\mathcal{I}$ indicates whether the corresponding point belongs to the task-relevant object instances. By concatenating the 3D attention map $\mathcal{I}$ with the 3D point cloud $\mathcal{P}$, we construct the state representation $z$ for the downstream visuomotor policy. Since the generated code selects one instance from all valid detected instances (e.g., all mugs), the computed 3D attention map helps resolve the language ambiguity.

\begin{figure*}[htbp]
    \centering
    \includegraphics[width=\linewidth]{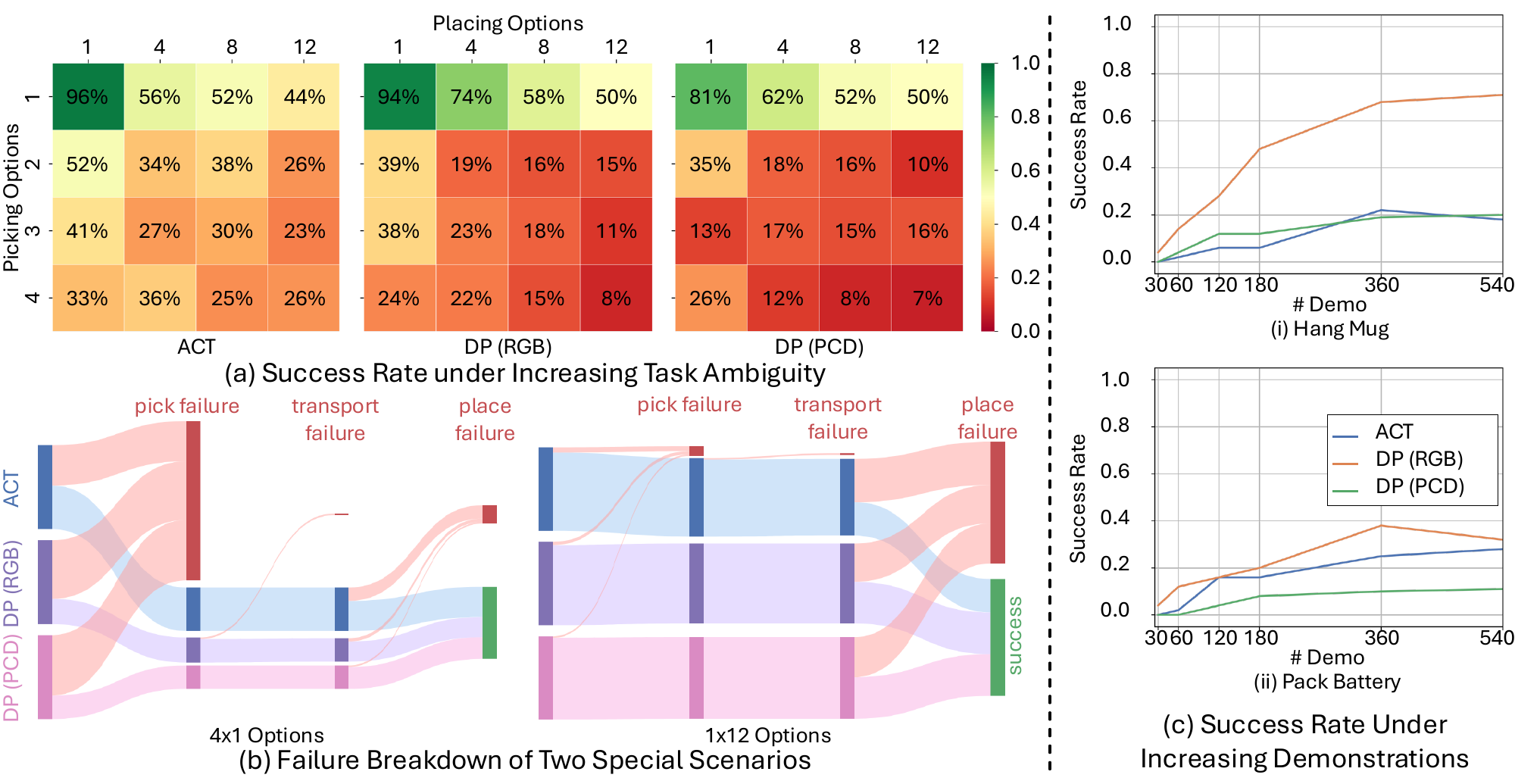}
    \vspace{-15pt}
    \caption{\small
    \textbf{Analysis of \yxadd{Existing} Imitation Learning Algorithms.}
    \textbf{(a)} We first evaluate three SOTA imitation learning algorithms--ACT, DP (RGB), and DP (PCD)--on the \texttt{Pack Battery} task with increasing task ambiguity. Each element in the matrix represents the method's success rate under the corresponding task setting, where columns indicate the number of empty slots and rows indicate the number of batteries to pick. \textit{A rollout is considered successful if a battery is placed into any slot.} All policies are trained on 180 demonstrations. For simple tasks with no ambiguity (top-left entry), the high success rate confirms the validity of the baseline methods. However, as the number of picking and placing options increases, success rates decline, highlighting the vulnerability of existing methods to task ambiguity.
    \textbf{(b)} We further analyze failure patterns in cases with more picking options (4×1, bottom-left entry of (a)) and more placing options (1×12, top-right entry of (a)) respectively. We observe that failure primarily occurs at the task stage with the highest ambiguity, demonstrating a strong correlation between policy failure and task ambiguity.
    \textbf{(c)} While (a) and (b) examine existing imitation learning algorithms trained on 180 demonstrations, we also investigate whether increasing the number of demonstrations mitigates the issue. For the \texttt{Pack Battery} task, we train and test on a mixture of 1 to 4 picking options with 1 placing option. For the \texttt{Hang Mug} task, we train and test on the scene with 2 mugs for picking and 3 branches for placing. We find that adding additional demonstrations in these settings often shows diminishing returns at low success rates even with extensive demonstrations, indicating that additional training data alone may not resolve the problem.
    }
    \label{fig:q1}
    \vspace{-15pt}
\end{figure*}

\subsection{Visuomotor Policy Learning}
\label{sec:policy}
Similar to Diffusion Policy~\cite{ho2020denoising, chi2023diffusionpolicy}, we model our policy as Denoising Diffusion Probabilistic Models (DDPMs). Instead of predicting the action directly, we train a noise prediction network $\epsilon_\theta$ conditioned on state representation $z$:
\begin{equation}
    \widehat{\epsilon^k} = \epsilon_\theta(a^k, z, k),
\end{equation}
where inputs are noisy actions $a^k$, current state representations $z=(\mathcal{P}, \mathcal{I})$, and denoising iterations $k$ and outputs are the noise $\widehat{\epsilon^k}$. During training, we sample denoising step $k$ and noise $\epsilon^k$ added to the unmodified sample $a^0$. Our training loss is defined as the Mean Square Error (MSE) between the added noise $\epsilon^k$ and the predicted noise:
\begin{equation}
    \mathcal{L} = \text{MSELoss}(\epsilon^k, \widehat{\epsilon^k}).
\end{equation}
At inference time, the policy begins with random actions $a^K$ and denoises them over $K$ iterations to generate the final action predictions. At each iteration, the action is updated according to the following equation:
\begin{equation}
    a^{k-1} = \alpha\bigl(a^{k}-\gamma\epsilon_\theta(a^{k},z,k)+\mathcal{N}(0,\sigma^2 I)\bigr),
\end{equation}
where $\alpha$, $\gamma$, and $\sigma$ are hyperparameters.

In contrast to the original variant of Diffusion Policy, where $z$ is the RGB observation, our $z$ is a state representation, concatenating 3D point clouds $\mathcal{P}$ and 3D attention map $\mathcal{I}$. Since the 3D attention map can highlight object instances relevant to the task instruction $l$, we adopt a 3D attention map as an intermediate representation to bridge a high-level VLM and a low-level visuomotor policy. Therefore, our framework can leverage a VLM's reasoning capability and generate low-level actions to accomplish the task following the language instruction.
We adopt PointNet++~\cite{qi2017pointnet++} to process point cloud inputs, with additional residual connection.

To train the policy, we collect the human demonstrations using a teleoperation interface in the real world or scripted policy in the simulation. In addition, we include a lightweight annotation process to label reference DINOv2 feature for reference objects and to generate ground-truth 3D attention maps. At inference time, our system predicts actions given the current observations, instruction, and a list of reference object features in the scene.

\section{Experiments}

In this section, we aim to answer four questions:  
(1)~How does the current imitation learning policy perform as task ambiguity increases, and can more data alone resolve this issue? (Section~\ref{sec:exp_1})  
(2)~Does the 3D attention map generated by VLM-generated code align with the language instruction? (Section~\ref{sec:exp_2})  
(3)~Is the 3D attention map a suitable representation for the downstream visuomotor policy to handle task ambiguity? (Section~\ref{sec:exp_3})  
(4)~How well does the entire system perform in both comprehensive simulation and real-world evaluations? (Section~\ref{sec:exp_4}.)

\subsection{Experiment Setup}

We use SAPIEN as the platform for large-scale simulation experiments~\cite{Xiang_2020_SAPIEN}. For real-world robot experiments, we use the ALOHA system for data collection and evaluation~\cite{zhao2023learning}, along with four RealSense cameras positioned around the workspace to capture multi-view RGB-D observations.

We consider three practical tasks: \texttt{Pack Battery}, \texttt{Hang Mug}, and \texttt{Stow Books}. These tasks involve potentially complex ambiguities, such as multiple picking and placing options in the battery packing task. In addition to collecting real-world demonstrations, we design a lightweight labeling process to generate language instructions and 3D attention maps for training the low-level policy.

For method evaluation, we report the task success rate, where the success criteria are determined by the information provided to the method. For methods conditioned on language or attention, we consider a rollout successful if the task is completed in the desired manner, such as successfully following the language instruction or picking the highlighted mug and placing it on the highlighted branch. For baseline imitation learning methods without additional conditioning, a rollout is considered successful if the base task is completed, regardless of the specific route or mode of execution (e.g., as long as a mug is placed on a branch).
We note that this distinction results in a stricter evaluation metric for methods conditioned on additional instructional information.%

\subsection{Analysis of Existing Imitation Learning Algorithm}
\label{sec:exp_1}

In this section, we aim to study how well current imitation learning algorithms can handle task ambiguities. Specifically, we consider two state-of-the-art methods, Action Chunking Transformer (ACT)~\cite{zhao2023learning} and Diffusion Policy (DP)~\cite{chi2023diffusionpolicy} in comprehensive simulation evaluations. For DP, we consider two variants - DP with RGB inputs, denoted as ``DP (RGB)'', and DP with point cloud inputs, denoted as ``DP (PCD)''. We note that here we do not consider language inputs and focus on studying the low-level policy capabilities.

In our first experiment, we evaluate ACT, DP (RGB), and DP (PCD) on the \texttt{Pack Battery} task with increasing task modality, as shown in Figure~\ref{fig:q1} (a). Each entry in the matrix in represents the method's success rate under the corresponding task setting, where columns indicate the number of empty slots to place the battery and rows indicate the number of batteries to pick. A rollout is considered successful if \emph{a} battery is placed into \emph{any} slot. All policies are trained on 180 demonstrations. For the simplest task (i.e., top-left entry), there is only one battery to pick and one fixed placing slot, meaning a single picking option and a single placing option. For the most ambiguous task (i.e., bottom-right entry), there are up to four batteries and twelve empty slots to choose from, corresponding to four picking options and twelve placing options.
Task-level ambiguity was gradually increased from the top left to the bottom right.

We observe that all baseline methods perform well in the simple task setting, demonstrating that the policy can successfully accomplish the task when no task ambiguity is present. However, as the task setting becomes more ambiguous due to an increasing number of possible choices, the performance of existing imitation learning algorithms degrades significantly. This suggests that current imitation learning algorithms struggle to handle task ambiguity.

Furthermore, we analyze the failure patterns of two specific cases--four picking options with one placing option (i.e., 4×1 options) and one picking option with twelve placing options (i.e., 1×12 options)--as shown in Figure~\ref{fig:q1} (b). As the number of picking options increases, ACT and DP struggle to execute the picking skill. Similarly, as the number of placement options increases, most failures occur during the placement stage of the task. The observed correlation between (i) increased task ambiguity and (ii) declining task success rates further underscores the limitations of existing imitation learning algorithms in handling task ambiguity.

In addition, we investigate whether increasing the number of demonstrations can help existing imitation learning algorithms handle task ambiguity. We conduct experiments on the \texttt{Pack Battery} and \texttt{Hang Mug} tasks in simulation. For the \texttt{Pack Battery} task, we train and test on a mixture of 1 to 4 picking options with 1 placing option. For the \texttt{Hang Mug} task, we train and test on a scene with 2 mugs for picking and 3 branches for placing. The number of demonstrations is increased from 30 to 540 episodes. While the performance of ACT and DP initially improves, they generally show diminishing returns while success rate is still low, and in some cases plateaus as the number of demonstrations further increases, suggesting that additional demonstrations may not effectively resolve task ambiguity.

\begin{figure*}[htbp]
    \centering
    \includegraphics[width=\linewidth]{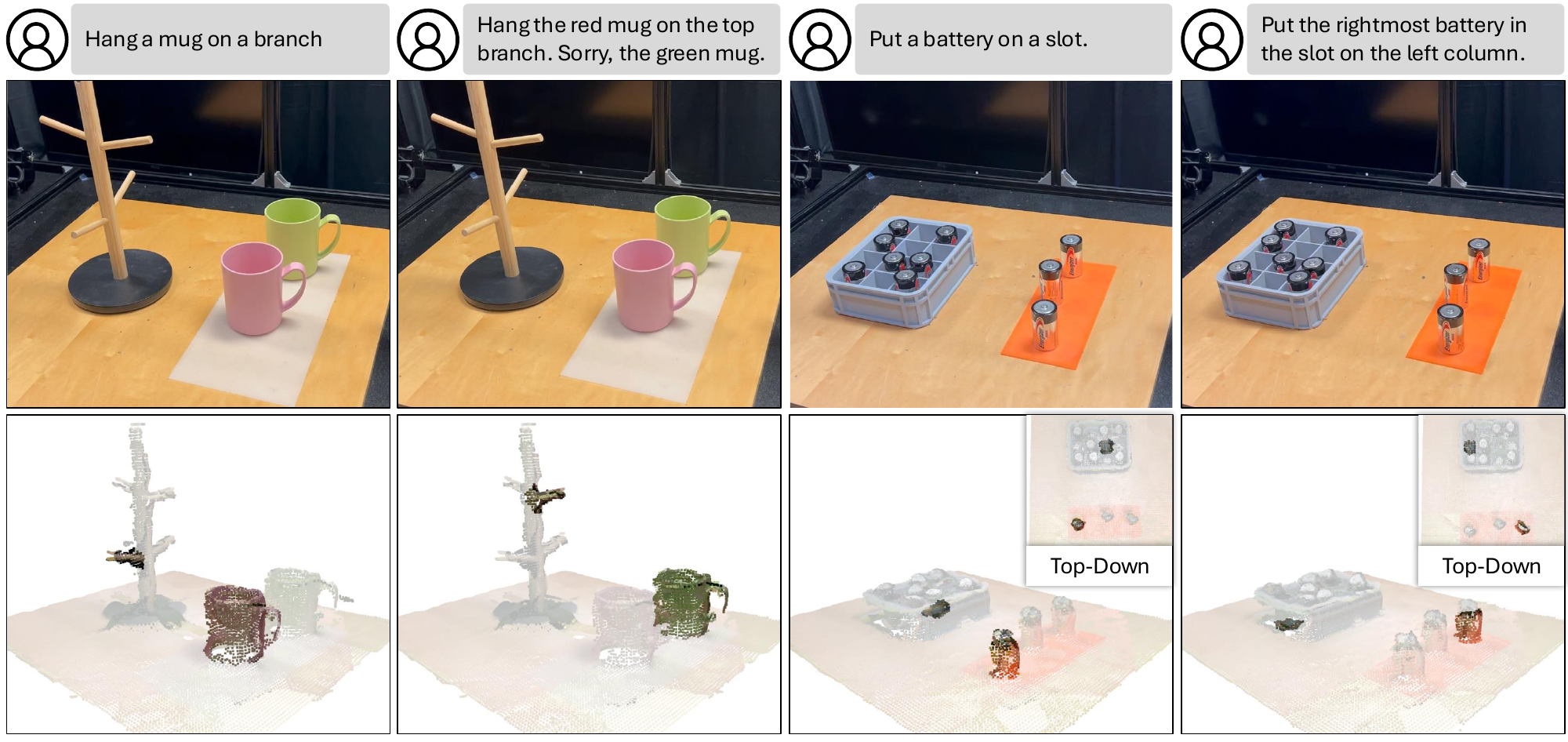}
    \vspace{-15pt}
    \caption{\small
    \textbf{3D Attention Maps Visualization.}
    We visualize the 3D attention maps for corresponding instructions and scenarios. First, our 3D attention maps successfully highlight the correct object instances even under ambiguous instructions, such as ``Hang a mug on a branch'' or ``Put a battery on a slot,''
    without specifying the instance. Furthermore, as instructions become more complex and/or specific, the 3D attention maps continue to attend to the correct instances, accurately matching the given instructions.
    }
    \vspace{-15pt}
    \label{fig:heatmap}
\end{figure*}

\subsection{Evaluation of 3D Attention Maps}
\label{sec:exp_2}

In this section, we evaluate the pipeline from language instructions to 3D attention maps across different scenes and instructions. Figure~\ref{fig:heatmap} illustrates our system's performance in various scenarios.
We observe that for a simple instruction such as ``Put the rightmost battery in the slot on the left column,'' the 3D attention maps correctly highlight the intended battery instance and slot position.
The VLM-generated code can also perform zero-shot interpretation of language exhibiting more complicated logical structures, such as self-repairing phrases such as ``Hang the red mug on the top branch. Sorry, the green mug.'' The generated 3D attention maps correctly highlight the green mug and top branch.
Furthermore, even with ambiguous commands like ``Hang a mug on a branch'' (without specifying a particular mug or branch), our system autonomously selects and highlights appropriate objects. These results demonstrate the system's ability to handle ambiguous or vague instructions while highlighting its semantic understanding and capability to generate accurate 3D attention maps across diverse instructions and scenarios.

\begin{table}[!ht]
\centering
\begin{tabular}{cccc}
\toprule
Scene & Hang Mug & Pack Battery & Total \\
\midrule
Ours & 97/100 & 94/100 & 191/200  \\
\bottomrule
\end{tabular}
\vspace{-5pt}
\caption{\small \textbf{Attention Quantitative Evaluation}. We quantitatively evaluate the pipeline from language instructions to 3D attention maps in simulation. Our results demonstrate that our pipeline effectively attends to task-relevant areas.}
\vspace{-10pt}
\label{tab:attn_quant}
\end{table}

In addition, we build a benchmark in the simulation to quantitatively evaluate the language-to-3D attention pipeline, which can automatically generate scenes, prompts, and corresponding ground truth 3D attention maps.
We measure the distance between ground truth 3D attention maps and generated 3D attention maps, and a test is considered successful if they are close enough.

Table~\ref{tab:attn_quant} summarizes our quantitative evaluation of the pipeline from language to attention. Our method effectively leverages the powerful visual-semantic understanding capabilities of VLMs and benefits from explicit spatial relation reasoning using 3D representations.
Additional analysis and visualizations of 3D attention failure cases are provided in the supplementary materials.

\subsection{Evaluation of Attention-Conditioned Diffusion Policy}
\label{sec:exp_3}

\begin{figure}
    \vspace{-10pt}
    \centering
    \includegraphics[width=\linewidth]{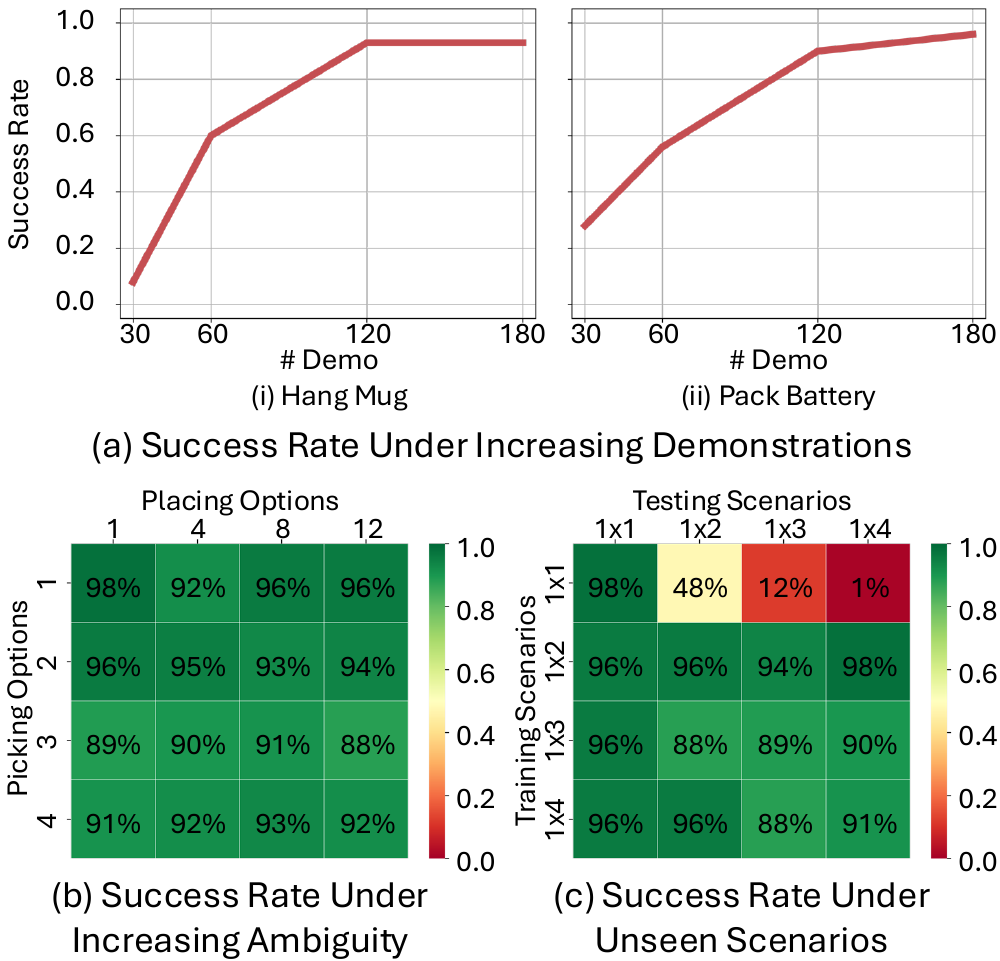}
    \vspace{-15pt}
    \caption{\small
    \textbf{Analysis of Attention-Conditioned Policy.}
    (a) We first examine the performance of the attention-conditioned policy under varying numbers of demonstrations. \textit{A rollout is considered as successful if the policy accomplishes the task as specified by the 3D attention maps}, a stricter criteria than that of Fig.~\ref{fig:q1}. The training and testing scenarios consist of a mixture of 1 to 4 picking options with 1 placing option. The success rate curve indicates that, given a sufficient number of demonstrations, our attention-conditioned policy converges to a high success rate.
    (b) We then analyze the performance of the attention-conditioned policy under increasing task ambiguity. Similar to Figure~\ref{fig:q1}, each element represents the success rate, with columns indicating the number of placing options and rows indicating the number of picking options. The success rate remains consistently high as task ambiguity increases, demonstrating that 3D attention serves as an effective representation for the downstream visuomotor policy.
    (c) Additionally, we find 3D attention improves policy generalization. When trained on scenarios with lower ambiguity, such as 1 picking option with 2 placing options (i.e., 1×2 in the row), the policy generalizes well to scenarios with greater ambiguity, such as those involving 3 or 4 placing options (i.e., 1×3 and 1×4 in the column).
    }
    \label{fig:q3}
    \vspace{-15pt}
\end{figure}

In this section, we investigate whether 3D attention is a suitable representation for visuomotor policy learning and evaluate the pipeline from 3D attention maps to low-level actions.
We first evaluate our method by varying the number of demonstrations on the \texttt{Pack Battery} task in simulation, as shown in Figure~\ref{fig:q3} (a). We train and test on a scene consisting of a mixture of 1 to 4 picking options and 1 placing option. A policy rollout is considered successful only if it completes the task as specified by the given groundtruth 3D attention map. Figure~\ref{fig:q3} (a) shows that our system’s success rate reaches ($>$90\%) at approximately 120 demos, indicating that our method can effectively handle task ambiguity.
Additionally, we increase task ambiguity and observe its effect on the success rate, as shown in Figure~\ref{fig:q3} (b). Each entry in the matrix represents the success rate under the corresponding task setting, where rows indicate the number of picking options and columns indicate the number of placing options. Figure~\ref{fig:q3} (b) demonstrates that our system is not significantly affected by task ambiguity and performs well across both simple and complex task settings. These results confirm that the 3D attention map is a robust representation for downstream visuomotor policy learning in ambiguous task scenarios.

Additionally, we stress-test our visuomotor policy on the \texttt{Pack Battery} task across unseen scenarios in simulation. In Figure~\ref{fig:q3} (c), each entry represents the success rate of the trained policy in different testing environments, where rows indicate training environments and columns indicate testing environments.  
For example, the second row (i.e., 1×2) corresponds to the training scenario with one picking option and two placing options, while the third column (i.e., 1×3) represents the testing scenario with one picking option and three placing options.  
First, we observe that the success rate along the diagonal is high, validating the expected performance pattern where policies perform well on their original training scenarios.
Second, while training on a 1x1 scenario does not generalize to scenarios with multiple placing options, the generalization of CodeDiffuser quickly improves after seeing more than one placing option at training time. The 3D attention module enables the policy to focus on task-relevant visual regions, thereby enhancing its ability to generalize across diverse and previously unseen task settings.

\subsection{Evaluation of the Entire System}
\label{sec:exp_4}

\begin{figure*}[htbp]
    \centering
    \includegraphics[width=\linewidth]{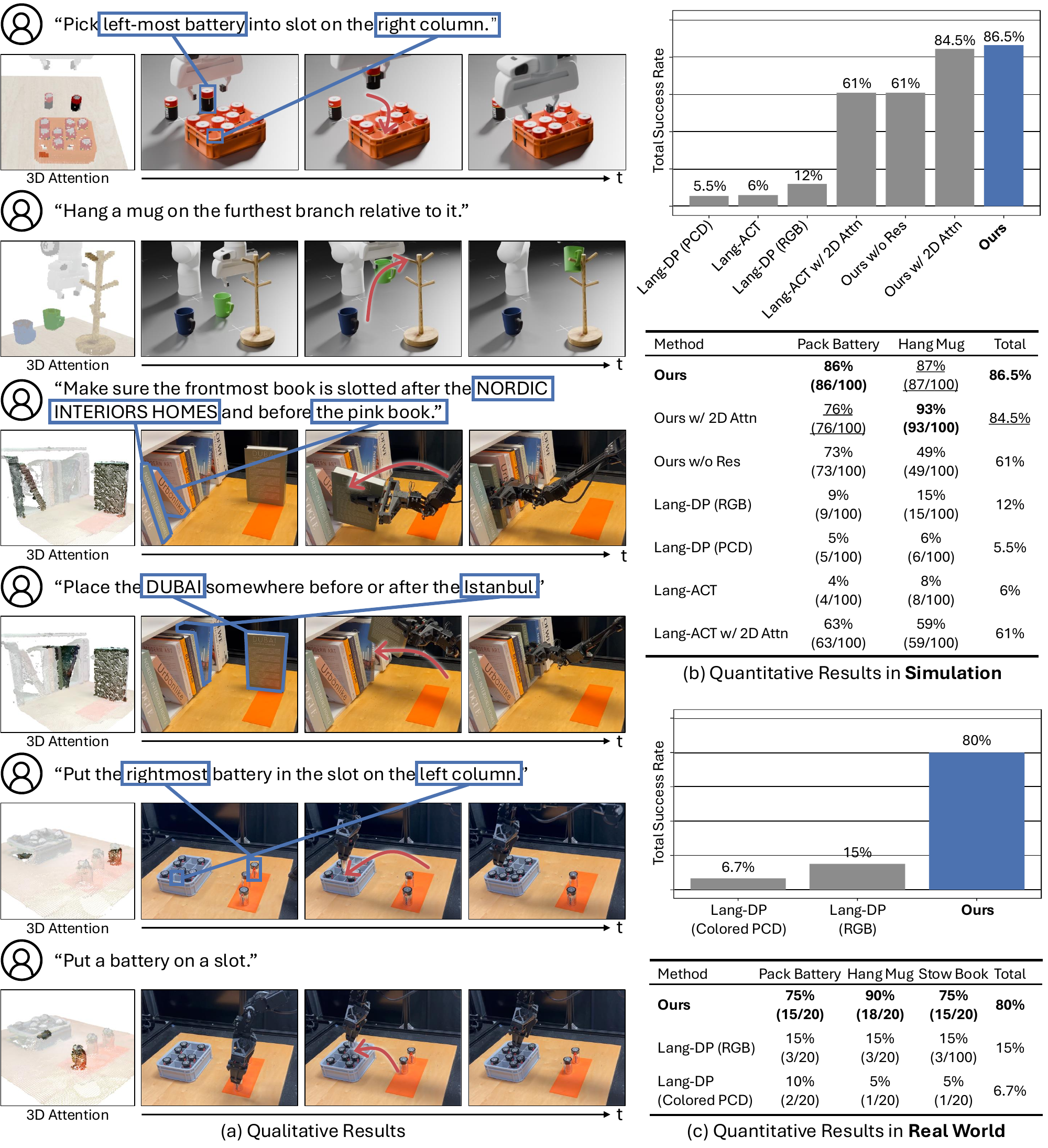}
    \vspace{-15pt}
    \caption{\small
    \textbf{Evaluation of Entire System.}
    (a) We qualitatively evaluate the entire pipeline, from language instructions to low-level actions, demonstrating how our system interprets semantic meanings from abstract instructions. Given similar initial configurations, our system can react appropriately to different instructions, showcasing its semantic understanding capability.
    (b) In the simulation, we compare our method against various baselines on both \texttt{Pack Battery} and \texttt{Hang Mug} tasks. A policy rollout is considered successful if it manipulates the object as specified by the instruction. Our results show that our system significantly outperforms baseline methods, highlighting its ability to comprehend high-level semantic information and execute complex tasks effectively.
    (c) We further validate our approach in real-world experiments using the same evaluation metric. Our system successfully performs challenging tasks involving contact-rich manipulation and multi-object interactions, consistently surpassing baseline methods in performance.
    }
    \label{fig:q4}
    \vspace{-15pt}
\end{figure*}

In this section, we evaluate the entire robotic manipulation framework, from language to low-level actions, through both quantitative and qualitative analyses. In the simulation, we collect 180 training demonstrations for each task. We consider two simulation tasks: \texttt{Hang Mug} and \texttt{Pack Battery}, which involve language ambiguity, contact-rich manipulation, and multi-object interactions. The \texttt{Hang Mug} task requires picking between two mugs and placing them on one of four branches, while the \texttt{Pack Battery} task involves picking from four batteries and placing them into one of twelve slots. For the simulation experiments, we compare our method against the following baselines:
\begin{itemize}
    \item \textbf{Ours with 2D Attention}: Instead of mapping the multi-view RGBD observation into 3D space, this baseline uses a 2D attention mechanism to segment objects of interest. The masked observation is then input into visuomotor policy.
    \item \textbf{Ours without Residual Connection}: In our policy, we include a residual connection in PointNet++ for visual feature extraction. This baseline ablates the residual connection to evaluate its contribution to performance.
    \item \textbf{Lang-DP (RGB)}: This baseline extends DP (RGB) by conditioning the policy on language using a frozen CLIP encoder. The extracted language features are concatenated with visual features to condition the diffusion policy.
    \item \textbf{Lang-DP (PCD)}: Similar to Lang-DP (RGB), this baseline adds language conditioning to DP (PCD) using a CLIP-based language encoder.
    \item \textbf{Lang-ACT}: This baseline augments the original ACT framework with language features, similar to Lang-DP.
    \item \textbf{Lang-ACT with 3D Attention}: Unlike the original ACT, which uses multi-view RGB observations, this baseline inputs 3D attention maps into Lang-ACT to assess whether the attention module consistently improves the performance of base imitation learning algorithms.
\end{itemize}
\begin{figure*}
    \centering
    \includegraphics[width=\linewidth]{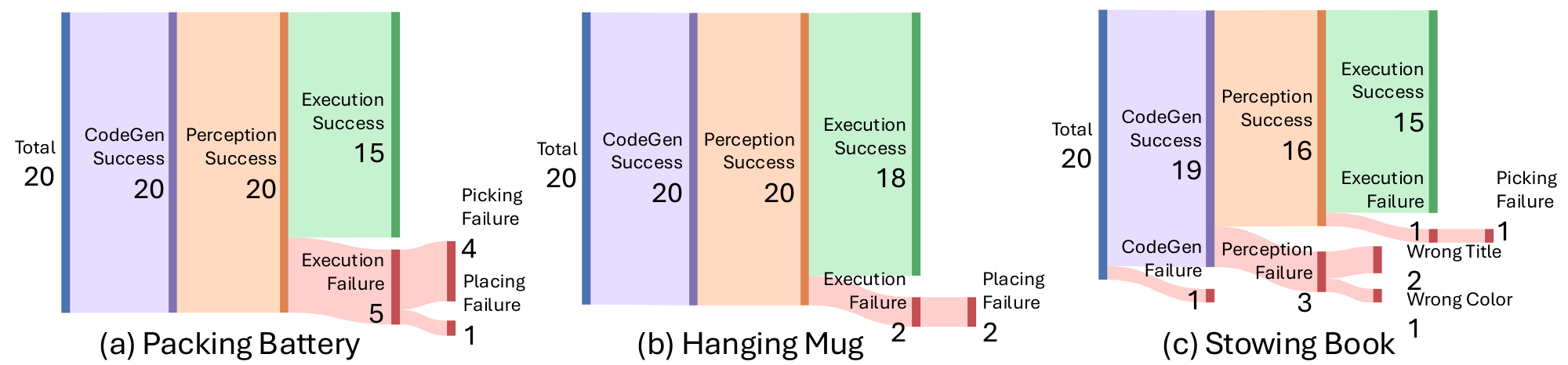}
    \vspace{-10pt}
    \caption{\small
    \textbf{System Failure Breakdown.} We categorize the failure patterns in the real-world experiments into code generation failures, perception failures, and execution failures. Our results indicate that the majority of failures occur during task execution, while code generation and perception remain relatively stable.
    }
    \label{fig:sys_failure}
    \vspace{-15pt}
\end{figure*}

The prompts are generated from randomly selected descriptive components, such as ``right," ``furthest," and ``blue." These components are incorporated into templates such as ``Put the blue mug into the furthest slot." More specifically, all prompts can be categorized into four types:
\begin{itemize}
\item \textbf{Prompt without slackness}: All objects are strictly specified, such as ``Hang the left-most mug on the top branch.''
\item \textbf{Prompt with full slackness}: No objects are strictly specified, such as ``Hang a mug on a branch.''
\item \textbf{Prompt with picked-object slackness}: The placement locations (e.g., branch or slot) are strictly specified, while the objects being placed are not, such as ``Hang a mug on the left-most branch.''
\item \textbf{Prompt with placed-object slackness}: The objects being placed (e.g., mug or battery) are strictly specified, while the placement locations are not, such as ``Hang the blue mug on a branch.''
\end{itemize}

Figure~\ref{fig:q4} (b) summarizes the quantitative results from the simulation experiments. From this table, we draw the following conclusions:
(1) Adding the attention module significantly enhances the policy's ability to accomplish tasks involving linguistic ambiguity. Compared to Lang-DP (PCD), our method leverages VLM to interpret language instructions and compute 3D attention maps, resulting in a substantial performance improvement from 5.5\% to 86.5\%.
(2) The 3D attention maps can also be integrated into other base imitation learning algorithms to improve performance. For instance, comparing Lang-ACT with 3D Attention to Lang-ACT, we observe a significant performance increase from 6\% to 61\%.
(3) The attention mechanism is also effective for 2D representations. In contrast to Lang-DP (RGB), our method, which incorporates a similar pipeline from language instructions to 2D attention, achieves a performance improvement from 12\% to 84.5\%.
(4) Our ablation study demonstrates that incorporating the residual connection into PointNet++ improves performance from 61\% to 86.5\%. This enhancement is attributed to the residual connection's ability to better propagate attention information into the  visuomotor policy, thereby improving trajectory prediction.

In addition, we evaluate the entire system on the \texttt{Hang Mug}, \texttt{Pack Battery}, and \texttt{Stow Book} tasks in the real world. We collect 150 demonstrations for each real-world task.
For the \texttt{Stow Book} task--a challenging task that is difficult to simulate due to its contact-rich nature--we test on scenes with two available slots for placement. We note that while using 2D attention achieves similar performance to using 3D attention maps, we adopt 3D attention maps for real-world tasks due to their robustness to environmental factors, as observed in DP3 and GenDP~\cite{Ze2024DP3,wanggendp}. We compare our method to Lang-DP (RGB) and Lang-DP (Colored PCD), where the point cloud is colored based on RGB observations.  

Figure~\ref{fig:q4} (c) summarizes the quantitative evaluation results from the real-world experiments. We find that our policy consistently outperforms the baselines by leveraging VLM-generated code as an interpretable and executable intermediate representation, effectively utilizing the visual-semantic reasoning capabilities of the VLM.

Figure~\ref{fig:q4} (a) presents our qualitative evaluation. Our policy effectively handles task instructions with varying degrees of specificity. For instance, the instruction ``Put a battery on a slot.'' is ambiguous regarding the target battery and slot. Our system interprets this ambiguous instruction, selects a specific instance, and successfully executes the task. In contrast, another instruction with a similar initial object configuration explicitly specifies the target battery and slot. Thanks to the VLM's visual-semantic reasoning capabilities, our system correctly interprets this precise instruction, identifies the appropriate task-relevant locations, and successfully completes the task. 

In addition, we analyze the common failure cases of our method, as shown in Figure~\ref{fig:sys_failure}. We break down the failure pattern according to the pipeline modules, including code generation failure, perception failure, and task execution failure. We observe that the majority of failure focuses on task execution, while the code generation and perception are relatively stable.

\section{Conclusion}

Language ambiguity is a common challenge in robotic manipulation, such as determining which mug to pick and where to place it for the instruction ``Hang a mug on a branch.'' Existing imitation learning algorithms typically employ end-to-end models that jointly interpret high-level semantic information and generate low-level actions, often resulting in suboptimal performance.  
In this work, we address this challenge by introducing a novel robotic manipulation framework that utilizes VLM-generated code as an executable and interpretable intermediate representation. The generated code interfaces with perception APIs to compute 3D attention maps using VFMs, which are then used for downstream visuomotor policy execution. Our modular design leverages both the visual-semantic understanding capabilities of VLMs and the smooth trajectory prediction of low-level policies.
In our experiments, we first identify the key limitations of existing imitation learning algorithms. We then conduct a comprehensive evaluation of our method in both simulation and real world, and study the pipeline from language to 3D attention maps, the pipeline from 3D attention maps to low-level actions, and the entire system respectively. We demonstrate CodeDiffuser's effectiveness in challenging robotic tasks involving task-level ambiguity, contact-rich 6-DoF manipulation, and multi-object interactions.

\textbf{Limitations:} The proposed method relies on VLMs and VFMs to generate codes, detect objects, and compute 3D attention maps. Therefore, the performance of our system is bounded by VLMs and VFMs. In the future, having more advanced VLMs and VFMs can benefit our system and make our system more robust. In addition, our perception APIs currently operate on an object level, which can pose challenges in generalizing to instructions involving deformable objects, such as ``grab the top of the circle of dough''. In addition, the additional manual annotation process is practically challenging for large-scale training. Scaling this method to larger datasets by simplifying the manual annotation process is a promising direction for future work.

\section*{Acknowledgement}

This work is partially supported by the Toyota Research Institute (TRI), the Sony Group Corporation, Google, and Dalus AI.
We greatly appreciate the NCSA for providing computing resources. This article solely reflects the opinions and conclusions of its authors and should not be interpreted as necessarily representing the official policies, either expressed or implied, of the sponsors.

\bibliographystyle{unsrtnat}
\bibliography{reference}

\clearpage

\end{document}


\maketitle

\newcommand{\toyitong}{{\textcolor{red}{@ Yitong}}}
\newcommand{\toguang}{{\textcolor{orange}{@ Guang}}}
\newcommand{\toyixuan}{{\textcolor{blue}{@ Yixuan}}}
\newcommand{\yx}[1]{{\textcolor{blue}{[Yixuan: #1]}}}
\newcommand{\yxadd}[1]{{\textcolor{blue}{#1}}}
\newcommand{\huan}[1]{{\textcolor{brown}{[Huan: #1]}}}
\newcommand{\huanadd}[1]{{\textcolor{brown}{#1}}}
\newcommand{\katliu}[1]{{\textcolor{magenta}{[Katliu: #1]}}}
\newcommand{\katliuadd}[1]{{\textcolor{magenta}{#1}}}
\newcommand{\yz}[1]{{\textcolor{cyan}{[Yunzhu: #1]}}}
\newcommand{\yzadd}[1]{{\textcolor{cyan}{#1}}}
\newcommand{\guang}[1]{{\textcolor{orange}{[Guang: #1]}}}
\newcommand{\guangadd}[1]{{\textcolor{orange}{#1}}}

\makeatletter
\def\blfootnote{\gdef\@thefnmark{}\@footnotetext}
\makeatother

\setlist[itemize]{left=0mm}

\blfootnote{$^*$Denotes equal contribution.}

\section{Method}
\subsection{Clustering and Segmentation}
For clustering, we first leverage DINO features to measure the similarity between the source feature and the features of point cloud. A predefined threshold is used to filter attention points based on the cosine similarity value. To ensure uniform point density across different parts, we apply Furthest Point Sampling (FPS) to downsample the point cloud to 8,000 points. Next, we employ a density-based clustering method on the selected attention points. Specifically, the clustering algorithm groups points into clusters based on a predefined L2-norm distance threshold—0.1 for packing a battery and 0.15 for hanging a mug.

We found DBSCAN performs robustly in our experiments, as evidenced by the 3D attention benchmark (Table~\ref{tab:attn_quant}) and the failure breakdown analysis (Main Figure 9). Additionally, due to modular design, it can be easily replaced with a more robust alternative as needed.

For methods utilizing 2D attention, we project the 3D point cloud with attention onto 2D images, generating a binary mask where attention regions are marked as 1 and non-attention regions as 0. Due to the sparse and fragmented nature of the initial projection, we apply dilation to refine the mask, ensuring a more cohesive and continuous attention region.

\subsection{Spatial Selection through CodeGen}
We utilize code as an interface to detect the target object based on spatial instructions, such as "the rightmost mug." Given a language instruction, the LLM generates the corresponding code and leverages NumPy APIs to analyze the spatial relationships between attention clusters, ultimately identifying the specified target.

For example, if the language instruction is ``\textit{Pick a battery into the slot on the middle columns.}'', the generated code would be:

\begin{lstlisting}
# compose API summary
# INSTRUCTION
# Pick a battery into the slot on the middle columns.

# RESPONSE
# Determine target object, battery
battery = self.detect("battery")[0]["pointer"]

# Determine target object, the slot on the middle column

# Detect slots
slot_list = self.detect("slot")
slot_centroid_list = []
for obj in slot_list:
    slot_centroid_list.append(np.mean(obj["pcd"], axis=0)[np.newaxis, :])
slot_centroid_numpy = np.concatenate(slot_centroid_list, axis=0)

# Extract x coordinate
slot_x = slot_centroid_numpy[:, 0] # use x to determine right or left

# The slot with the max x value and the min x value
slot_max_x = np.max(slot_x)
slot_min_x = np.min(slot_x)

# Determine the x coordinate range for the middle column
middle_x_threshold = (slot_max_x + slot_min_x) / 2

# Find the index of the slot closest to the middle_x_threshold
tgt_slot_idx = np.argmin(np.abs(slot_x - middle_x_threshold))

# Extract the points of the target slot
tgt_slot = slot_list[tgt_slot_idx]["pointer"]

output_var = [battery, tgt_slot]
# -------------
# detection API summary
# PROMPT
# ['battery', 'slot']
# battery

# RESPONSE
battery_list = self.get_obj('battery')
output_var = battery_list
# -------------
# detection API summary
# PROMPT
# ['battery', 'slot']
# slot

# RESPONSE
slot_list = self.get_obj('slot')
output_var = slot_list
\end{lstlisting}

\subsection{Visual Selection through VLM}
We employ a Vision-Language Model (VLM) to detect the target object based on visual instructions such as "the orange book." When visual detection is required, the generated code can invoke a predefined API, \textit{find instance in category}, which annotates instances in RGB images with corresponding labels. The VLM then interprets the language instruction and provides the index of the matching label as shown in Main Figure~4.
For example, if the language instruction is ``\textit{I want to use the blue mug to drink some water. Put away the other one on a branch.}'', the generated code would be:

\begin{lstlisting}
# compose API summary
# INSTRUCTION
# I want to use the blue mug to drink some water. Put away the other one on a branch.

# RESPONSE
# determine target object, the other mug (not blue)
blue_mug = self.detect("blue mug")[0]
mug_list = self.detect("mug")
tgt_mug_list = []
for mug in mug_list:
  if mug != blue_mug:
    tgt_mug_list.append(mug)
tgt_mug = tgt_mug_list[0]["pointer"]

# determine target object, branch
branch = self.detect("branch")[0]["pointer"]

output_var = [tgt_mug, branch]
# -------------
# detection API summary
# PROMPT
# ['mug', 'branch']
# blue mug

# RESPONSE
mug_list = self.get_obj('mug')
tgt_idx = self.find_instance_in_category(instance = 'blue mug', category = 'mug')
tgt_mug = []

for i in tgt_idx:
    tgt_mug.append(mug_list[i])
output_var = tgt_mug
# -------------
# find_instance_in_category API summary
# PROMPT
# blue mug


# RESPONSE
# The object labeled with index 1 is a blue mug.
1
# -------------
# detection API summary
# PROMPT
# ['mug', 'branch']
# mug

# RESPONSE
mug_list = self.get_obj('mug')
output_var = mug_list
# -------------
# detection API summary
# PROMPT
# ['mug', 'branch']
# branch

# RESPONSE
branch_list = self.get_obj('branch')
output_var = branch_list
\end{lstlisting}

\section{Additional 3D Attention Evaluation}

In addition, we build a benchmark in the simulation to quantitatively evaluate the language-to-3D attention pipeline, which
can automatically generate scenes, prompts, and corresponding
ground truth 3D attention maps. We measure the distance
between ground truth 3D attention maps and generated 3D
attention maps, and a test is considered successful if they
are close enough.
Specifically, we use Chamfer Distance to evaluate alignment between GT and generated 3D attention. A threshold of 0.005 is used to decide spatial proximity.
In this experiment, we compare with the
following two baselines:
\begin{itemize}
    \item G-SAM2: We directly prompt Grounded-SAM2 to segment
out the object to attend to.
    \item G-SAM2+GPT-4o: We first segment out objects for each
object category using Grounded-SAM2 and then ask GPT4o to select the object instance.
\end{itemize}

\begin{table}[!ht]
\centering
\begin{tabular}{cccc}
\toprule
Scene & Ours & G-SAM2 & G-SAM2+GPT-4o \\
\midrule
Hang Mug & \textbf{97/100} & 0/100 & 0/100  \\
Pack Battery & \textbf{94/100} & 0/100 & 0/100 \\
\midrule
Total & \textbf{191/200} & 0/100 & 0/100 \\
\bottomrule
\end{tabular}
\vspace{-5pt}
\caption{\small \textbf{Additonal Attention Quantitative Evaluation}. We quantitatively evaluate the pipeline from language instructions to 3D attention maps in simulation. Our results demonstrate that our pipeline effectively attends to task-relevant areas, whereas baseline methods fail due to their limited fine-grained visual-semantic reasoning and spatial understanding capabilities.}
\vspace{-10pt}
\label{tab:attn_quant}
\end{table}

\begin{figure}
  \centering
  \includegraphics[width=\linewidth]{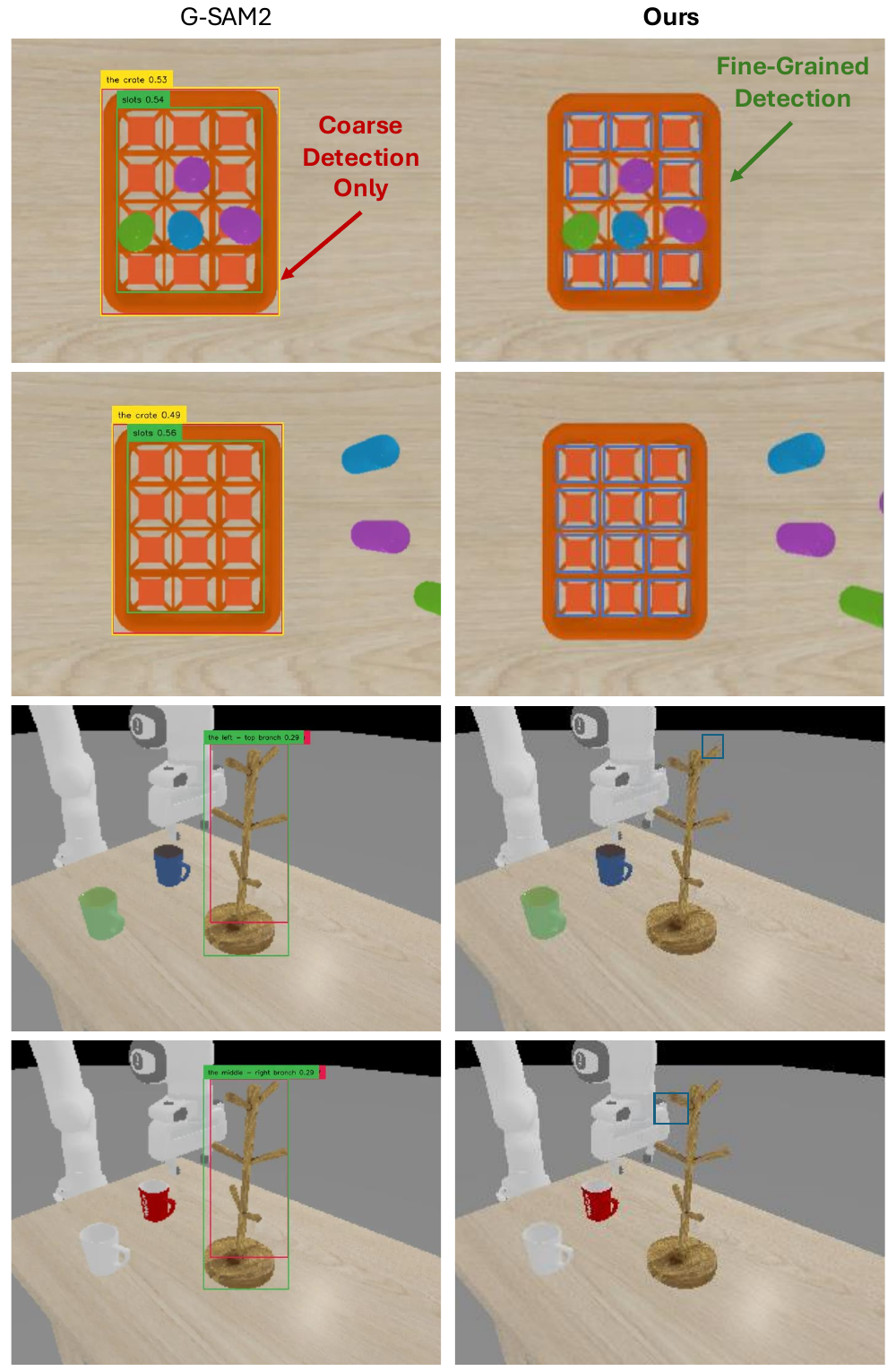}
  \caption{\small
  \textbf{Comparisons with Baselines on 3D Attention.} Compared to G-SAM2, our method allows for more fine-grained visual understanding, such as detecting slots and branches.
  }
  \label{fig:supp_attn_comp}
\end{figure}

Table~\ref{tab:attn_quant} summarizes our quantitative evaluation of the
pipeline from language to attention. We find that our method
outperforms the baselines, as it leverages the powerful visualsemantic understanding capabilities of VLMs and benefits
from explicit spatial relation reasoning using 3D representations. In contrast, Grounded-SAM2 lacks the capability for
fine-grained semantic reasoning, such as detecting branches
and slot locations, resulting in zero success.
Qualitative comparisons between our method and G-SAM2 are shown in Figure~\ref{fig:supp_attn_comp}.

\section{Simulation Experiments}

\subsection{Training Data Generation and Labeling}

In this project, we conduct experiments in Sapien using two scenarios: packing batteries and hanging mugs. We use scripted policies to generate demonstrations. During data generation, after defining the task scenes, we randomly select the target objects (e.g., a battery and slot, or a mug and branch) and generate the corresponding manipulation trajectories. At the same time, we log the ground truth attention locations for future labeling. This includes details such as which object to manipulate and where to place it.

For the labeling process, we use DINO features to query the attention point cloud, cluster the points into distinct parts, and identify the target part based on the ground truth location information. For methods involving 2D attention, we generate a point cloud with attention points and project it onto images as segmentation masks.

\subsection{Evaluation Benchmark with Language Instruction}

For system evaluation, the main objective is to assess whether the policy can accurately perform tasks based on language instructions. The system’s inputs for evaluation include visual observations and language prompts. Specifically, the visual observations are RGBD images captured from five different viewpoints, while the prompts are generated based on the task scene. These prompts are then validated using a rule-based approach to ensure their feasibility and coherence. Additionally, we define success criteria derived from the given prompts and task scenes. To be considered successful, the system must meet these criteria within a predefined time limit (e.g., 240 frames).

The scenario generation process can be broken down into several parts.
\begin{itemize}
\item \textbf{Descriptive Components Selection and Instruction Generation}:
Descriptive components (e.g., “blue mug,” “furthest branch” for hanging a mug, “slot in the first row,” “left battery” for packing a battery) are randomly sampled from a predefined library. Once the descriptive components are selected, we randomly choose a template from an instruction library and combine it to form a complete language instruction. A more detailed explanation can be found in the next subsection.

\item \textbf{Task Scene Randomization and Matching}:
Using the selected descriptive components, we generate task scenes in Sapiens (e.g., initial position, pose, number, and color of batteries). It is crucial that the correlation between the physical scenes and the prompts is accurate. For example, if the prompt is “hang a blue mug,” there should be at least one blue mug on the table. Similarly, if the prompt is “put the battery into the slot in the first row,” at least one slot in the first row must be available. We use a predefined check function, along with the descriptive components, to ensure that the generated scene matches the instructions. If the scene does not meet the requirements, a new one is generated and checked again. This process continues until the generated scene satisfies all conditions.

\item \textbf{Success Criteria Specification}:
We can also obtain the indices of objects using a predefined extraction function, which works in conjunction with the descriptive components. This function is designed to identify target objects. For example, the \textit{blue\_mug\_extract} function retrieves the indices of all blue mugs on the table. By using this function, we can determine the positions of the target objects referenced in the instructions and get access to their poses through API defined in Sapiens.
\end{itemize}

In this way, we can automatically evaluate whether the policy executes successfully based on our language instructions, which are also generated without manual effort. The detailed criteria are listed below:

\begin{itemize}
    \item \textbf{Pack Batteries}: The reward function determines success based on three criteria: the object’s horizontal distance from the target must be less than 0.03 units, the object must be aligned with the global Z-axis (i.e., nearly upright) with a product value greater than 0.99, and the object's height must be less than 0.009 units. If all these conditions are met for any object-target pair, the function returns a reward of 1.0, indicating success; otherwise, it returns 0.0, indicating failure.
    \item \textbf{Hang Mugs}: The reward function evaluates success based on the position of an object relative to a target. Specifically, the object's X-coordinate must be within 0.05 units of the target branch's X-coordinate, and the object's Z-coordinate must be between the target branch's Z-coordinate and a lower offset of 0.1 units below it. If both conditions are met for any object-target pair, the function returns a reward of 1.0, indicating success. If no pair satisfies these criteria, the function returns 0.0, indicating failure.
\end{itemize}

The object-target pairs are determined based on the language instructions, specifying which pairs meet the requirements of the given instruction. The process for this determination is explained in \textbf{Success Criteria Specification}.

\subsection{Details of Language Instruction Generation}

The first step in generating language instructions is sampling descriptive components. These are object-centric descriptions, such as ``blue mug'', `` front-most battery'', ``furthest branch''. 
We consider both spatial information (e.g., ``furthest'', ``right'') and visual information (e.g., ``red'', ``white'') when selecting these components. The libraries for hanging mugs and packing batteries are listed below:

\begin{itemize}
    \item \textbf{Hang Mugs}: \textit{left-most mug}, \textit{right-most mug}, \textit{white mug}, \textit{red mug}, \textit{blue mug}, \textit{green mug}, \textit{furthest branch}, \textit{right-topmost branch}, \textit{left-topmost branch} and \textit{right-middle branch}.
    \item \textbf{Pack Batteries}: \textit{left-most battery}, \textit{right-most battery}, \textit{front-most battery}, \textit{back-most battery}, \textit{furthest slot}, \textit{nearest slot}, \textit{slot on the front-most row}, \textit{slot on the middle row}, \textit{slot on the back-most row}, \textit{slot on left column}, \textit{slot on the middle columns} and \textit{slot on the right column}.
\end{itemize}

Once the descriptive components are determined, we randomly select a template from an instruction library and combine it to form a complete language instruction. Below is the full list of templates for hanging mugs:

\begin{itemize}
    \item \textbf{No Slackness}:
    
        \textit{"Hang the \{mug\} on the \{branch\}."}
        
        \textit{"I want to use the \{other\_mug\} to drink some water. Put away the other one on the \{branch\}."}
        
        \textit{"I want to use the \{other\_mug\} to drink some water. Put away the other mug on the \{branch\}."}
        
        \textit{"I will use the \{other\_mug\} to drink some water. Hang the \{mug\} on the \{branch\}."}
        
        \textit{"Hang the \{mug\} on the \{other\_branch\}. Sorry, the \{branch\}."}
        
        \textit{"There are two mugs. Keep \{other\_mug\} on the table, put the other one on \{branch\}."}
        
        \textit{"I will not use this \{mug\} now. Hang it on \{branch\}"}
        
        \textit{"Put Bob's mug, the \{mug\}, on the \{branch\}."}

    \item \textbf{Mug Slackness}:
    
        \textit{"Hang a mug on the \{branch\}."}
        
        \textit{"Put away a mug on the \{branch\}."}
        
        \textit{"Hang a mug on the \{other\_branch\}. Sorry, the \{branch\}."}

    \item \textbf{Branch Slackness}:

        \textit{"Hang the \{mug\} on a branch."}
        
        \textit{"I want to use the \{other\_mug\} to drink some water. Put away the other one on a branch."}
        
        \textit{"I want to use the \{other\_mug\} to drink some water. Put away the other mug on a branch."}
        
        \textit{"I will use the \{other\_mug\} to drink some water. Hang the \{mug\} on a branch."}
        
        \textit{"Hang the \{other\_mug\} on a branch. Sorry, the \{mug\}."}
        
        \textit{"There are two mugs. Keep \{other\_mug\} on the table, put the other one on a branch."}
        
        \textit{"I will not use this \{mug\} now. Hang it on a branch"}
        
        \textit{"Put Bob's mug, the \{mug\}, on the a branch."}

    \item \textbf{Both Slackness}:
    
        \textit{"Hang a mug on a branch."}
        
        \textit{"There are two mugs. Keep one on the table, put the other one on a branch."}
    
\end{itemize}

This is the full list of template for packing batteries:

\begin{itemize}
    \item \textbf{No Slackness}:
    
        \textit{"Pick the \{battery\} outside the crate into the \{slot\}."}
        
        \textit{"I need to use the \{other\_battery\}. Put away the \{battery\} outside the box in the \{slot\}."}
        
        \textit{"The desk is too messy. Put away the \{battery\} outside the box into the \{slot\}."}
        
        \textit{"I want to put the \{battery\} outside the crate into the \{other\_slot\}, oh sorry, the \{slot\}."}
        
        \textit{"You should make the table more tidy, Just start from putting the \{battery\} outside the crate into the \{slot\}."}

    \item \textbf{Battery Slackness}:
    
        \textit{"Pick a battery outside the crate into the \{slot\}."}
        
        \textit{"The desk is too messy. Put away a battery outside the crate into the \{slot\}."}
        
        \textit{"I want to put a battery outside the crate into the \{other\_slot\}, oh sorry, the \{slot\}."}
        
        \textit{"I want to put a battery outside the crate into the \{slot\}."}

    \item \textbf{Slot Slackness}:

        \textit{"Pick the \{battery\} outside the crate into a slot."}
        
        \textit{"I need to use the \{other\_battery\}. Put away the \{battery\} outside the crate in a slot."}
        
        \textit{"The desk is too messy. Put away the \{battery\} outside the crate into a slot."}
        
        \textit{"I want to put the \{other\_battery\} into a slot, oh sorry, the \{battery\} outside the crate."}
        
        \textit{"You should make the table more tidy, Just start from putting the \{battery\} outside the crate into a slot."}

    \item \textbf{Both Slackness}:
    
        \textit{"Put a battery outside the crate on a slot."}
        
        \textit{"There are some batteries outside the crate. Put one into a slot."}
\end{itemize}

\subsection{Baseline Details}

\begin{itemize}
    \item \textbf{Ours with 2D Attention}: Instead of mapping the multi-view RGBD observation into 3D space, this baseline uses a 2D attention mechanism to segment objects of interest. The masked observation is then input into visuomotor policy. Specifically, we first obtain 3d point cloud from images and ground attention on it through our proposed pipeline. Then the 3d attention is projected onto 2d images in the format of segmentation mask, where attention region is 1 and none-attention region is 0. We use attention masks to multiply corresponding RGB images. We train diffusion policy based on this observation. The total training epoch number is 360 and the learning rate scheduler is cosine with 600 epochs maximum limitation. The training batchsize is 64.
    \item \textbf{Ours without Residual Connection}: In our policy, we include a residual connection in PointNet++ for visual feature extraction. This baseline ablates the residual connection to evaluate its contribution to performance. We train diffusion policy based on typical 3d point cloud with 3d attention as ours. The total training epoch number is 360 and the learning rate scheduler is cosine with 600 epochs maximum limitation. The training batchsize is 64.
    \item \textbf{Lang-DP (RGB)}: This baseline extends DP (RGB) by conditioning the policy on language using a frozen CLIP encoder. The extracted language features are concatenated with visual features to condition the diffusion policy. The initial embedding dimension is 1536, which is reduced into 128 after processed through a linear layer. We train diffusion policy based on RGB images and textual embedding. The total training epoch number is 360 and the learning rate scheduler is cosine with 600 epochs maximum limitation. The training batchsize is 64.
    \item \textbf{Lang-DP (PCD)}: Similar to Lang-DP (RGB), this baseline adds language conditioning to DP (PCD) using a CLIP-based language encoder. The initial embedding dimension is 1536, which is reduced into 128 after processed through a linear layer. We train diffusion policy based on 3d point cloud without attention, but with RGB channel. RGB information should be accessible for language reasoning, especially for visual reasoning such as ``blue mug''. The total training epoch number is 360 and the learning rate scheduler is cosine with 600 epochs maximum limitation. The training batchsize is 64.
    \item \textbf{Lang-ACT}: This baseline augments the vanilla ACT framework with language features, similar to Lang-DP. For language embedding, the initial embedding dimension is 1536, which is reduced into 128 after processed through a linear layer. We regard it as an additional token, which are processed similarly as RGB image tokens. The total training epoch number is 540, where we find ACT will show better performance with a bigger epochs number. For fair comparison, we take the advantages of all methods. The learning rate scheduler is constant. The training batchsize is 64.
    \item \textbf{Lang-ACT with 3D Attention}: Unlike the vanilla ACT, which uses multi-view RGB observations, this baseline inputs 3D attention maps into Lang-ACT to assess whether the attention module consistently improves the performance of base imitation learning algorithms.
    The total training epoch number is 540, where we find ACT will show better performance with a bigger epochs number and the learning rate scheduler is constant. The training batchsize is 64.
\end{itemize}

\subsection{Single vs. Recursive VLM Calls}

While we can use a single VLM call, we choose the recursive approach to leverage the modular design for problem decomposition.
To ablate this design choice, we evaluated the generated 3D attention map by single VLM call, following the experiment in Section~IV.C and Table~\ref{tab:attn_quant}. We observed that the success rate drops from 94\% and 97\% to 90\% and 91\%. We will add this to the revised paper. Second, recursive calls are also standard in existing works~\cite{huang2023voxposer}.

\begin{figure}
  \centering
  \includegraphics[width=\linewidth]{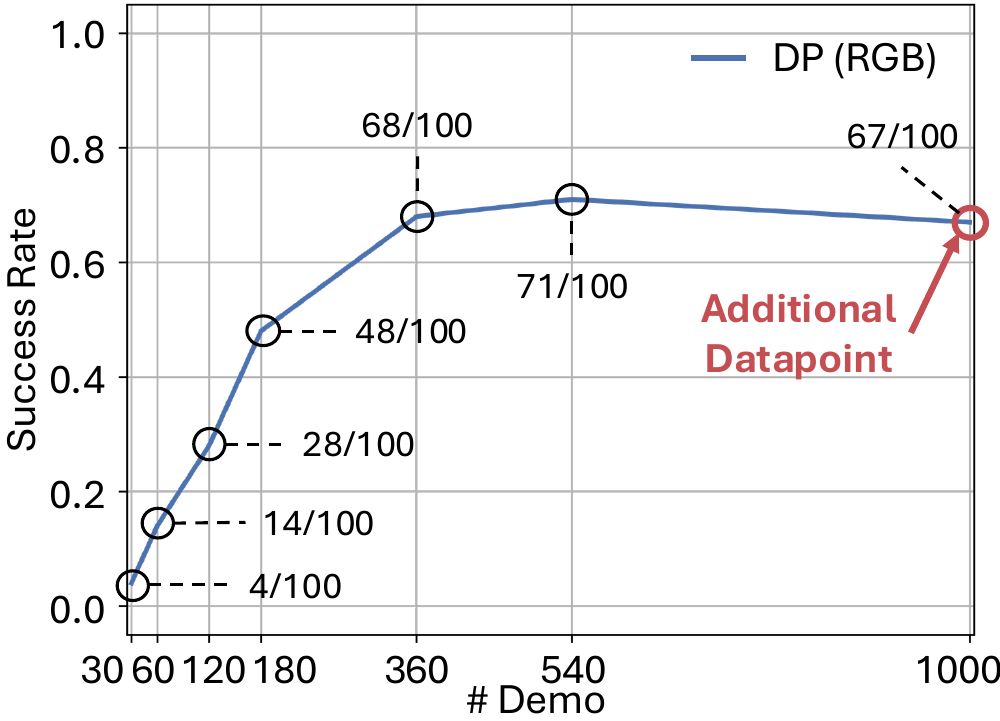}
  \caption{\small
  \textbf{Scaling Plot.} We increase the number of demonstrations to 1000 and observe the performance plateaus.
  }
  \label{fig:scale}
\end{figure}

\subsection{Scaling Plot}
To further demonstrate that the scaling plot (Fig. 4) for DP (RGB) is saturated, we increased the number of demonstrations to 1,000 and performed the same evaluation. The performance plateaued, confirming saturation (Fig.~\ref{fig:scale}).

\section{Real-World Experiments}

\begin{figure*}
    \centering
    \includegraphics[width=\linewidth]{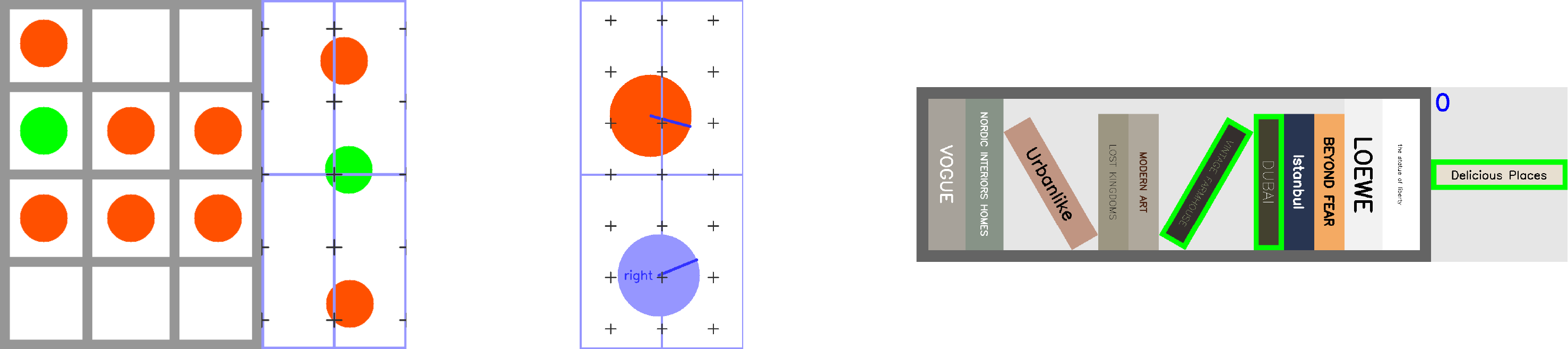}
    \caption{\small
    \textbf{Initial Configuration Visualization.} Initial configuration examples are generated by a program for real experiments. The leftmost visualization image corresponds to the packing a battery task, where orange circles represent batteries, and green circles indicate the target battery and target slot. The middle image represents the hanging a mug task. Circles of different colors correspond to mugs of the same color, while blue lines within the circles indicate the handle angles of each mug. The label "Right" signifies that the specified mug should be picked up and placed on the right branch. The cross markers on the pad image have exact corresponding locations to the cross markers on the real pad. This facilitates the operator in quickly identifying the relative positions of objects. Finally, the rightmost image corresponds to the stowing a book task, where books with green boundaries indicate the target book to be picked and the designated slot for placement.
    }
    \label{fig:init_config}
\end{figure*}

\subsection{Training Data Generation}

For the real-world experiment, datasets are collected, and policies are evaluated on Aloha~\cite{zhao2023learning} for three different tasks: packing a battery, hanging a mug, and stowing a book. The same dataset generation logic used in training applies to real experiments, but with several key differences. In the real experiment setup, there are five cameras: four positioned at each top corner and one at the center on top.

For data generation, the first step is to sample the required number of initial configurations for training demonstrations. According to predefined randomness—for example, the mug's position follows a uniform distribution within a fixed area—a program generates two output files: (1) a visualization image that helps the operator reset the scene before and after each demonstration as shown in Figure~\ref{fig:init_config}, and (2) a YAML script that describes the scene in a structured format, which is used for instruction generation. The same program is used to generate initial configurations for evaluation. During instruction generation, the instruction generator produces instructions based on the YAML scene description file and an instruction template file.

\begin{figure*}
    \centering
    \includegraphics[width=\linewidth]{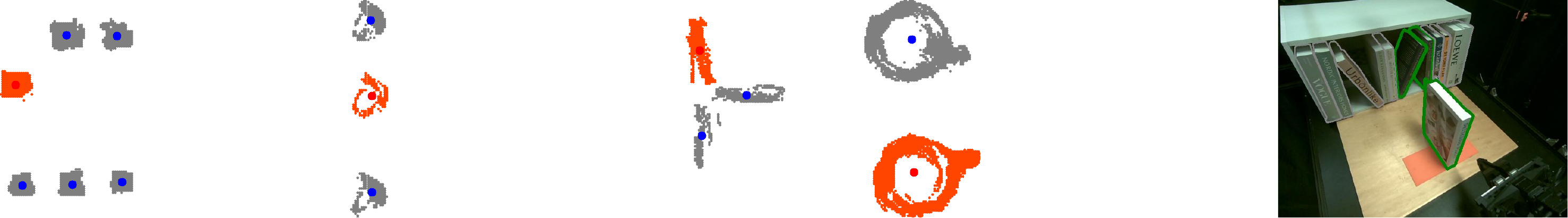}
    \caption{\small
    \textbf{Dataset labeling interface.} The left interface corresponds to the packing a battery task, while the middle interface is for the hanging a mug task. In both tasks, the target objects are highlighted in orange. The right interface represents the stowing a book task, where the target objects are highlighted with green boundaries. The operator only needs to move the cursor close to the target object and click to select it. Clicking the selected object again will deselect it.
    }
    \label{fig:labeling}
\end{figure*}

For labeling, we use an intuitive GUI labeling tool that allows the operator to quickly select objects with attention. The labeling tool for the packing battery and hanging mug tasks projects the point cloud of the entire scene in a top-down view. However, this approach is not suitable for the stowing book task because the books are placed too close together, making the top-down view cluttered. Instead, we project the point cloud from the perspective of one of the cameras, allowing the operator to directly select objects within the camera’s view. The labeling interfaces for each task are shown in Figure~\ref{fig:labeling}

\begin{figure*}
    \centering
    \includegraphics[width=\linewidth]{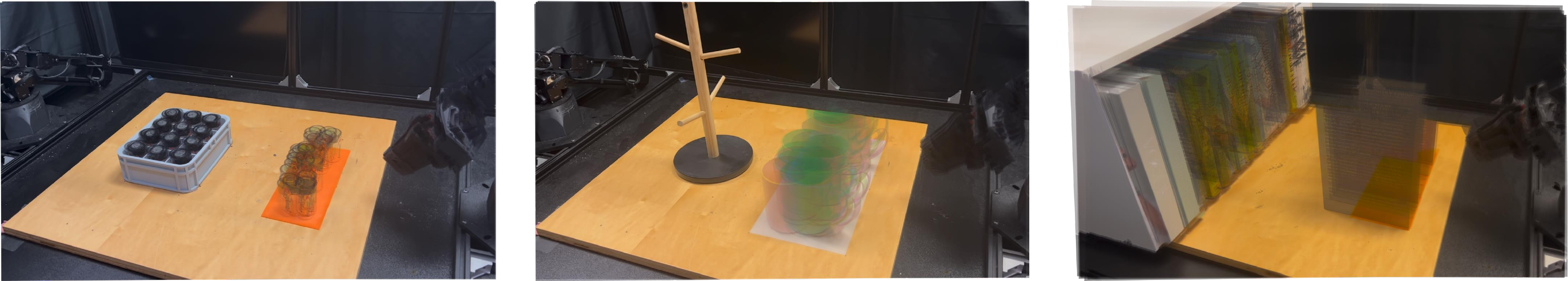}
    \caption{\small
    \textbf{Initial Configuration Overlay.} The randomness for each task is visualized as an overlay of initial configurations.
    }
    \label{fig:init_overlay}
    \vspace{-15pt}
\end{figure*}

\subsection{Task Design}
We carefully design each task to be complex enough to evaluate the effectiveness of our pipeline while ensuring an appropriate level of randomness, allowing the low-level policy to converge with a limited number of human demonstrations. The randomness visualization is shown in Figure~\ref{fig:init_overlay}.

\paragraph{Packing a Battery}  
In this task, a crate with 12 slots is arranged in 3 rows and 4 columns. In front of the crate, a random number of batteries, ranging from 1 to 3, are randomly placed on an orange pad. Additionally, some batteries may already be positioned inside the crate. The goal of the task is to place a battery on the pad into a slot according to a given linguistic instruction.

We define four types of instructions based on the level of specificity: \textbf{no slackness, battery slackness, slot slackness}, and \textbf{both slackness}.

\begin{itemize}
    \item \textbf{No Slackness:} Both the battery and the slot are explicitly specified. For example, \textit{``Put the leftmost battery into the slot in the back row.''} The task is only considered successful if the specified battery is placed into the designated slot.
    
    \item \textbf{Battery Slackness:} Only the slot is explicitly specified. For example, \textit{``Place a battery into the slot in the right column.''} Any battery can be selected, but it must be placed in the designated slot for the task to be considered successful.
    
    \item \textbf{Slot Slackness:} Only the battery is explicitly specified. For example, \textit{``I need to use the leftmost battery. Put away the middle battery in a slot.''} The specified battery must be picked up, but it can be placed in any available slot for the task to be considered successful.
    
    \item \textbf{Both Slackness:} Neither the battery nor the slot is explicitly specified. For example, \textit{``There are some batteries. Put one into a slot.''} In this case, placing any battery into any slot is considered successful.
\end{itemize}

The instruction types are randomly selected, and the instructions are generated using the instruction generator mentioned earlier.

\paragraph{Hanging a Mug}  
The experiment scene consists of a mug tree with four branches (the rear branch is not used) and two mugs randomly placed on a white pad. The angle of each mug handle ranges from \(-30^\circ\) to \(30^\circ\). Each mug has a color randomly selected from red, blue, or green, ensuring that no two mugs on the pad share the same color. The goal of this task is to pick up a mug and place it on a branch according to a given linguistic instruction.

We define four types of instructions based on the level of specificity: \textbf{no slackness, mug slackness, branch slackness}, and \textbf{both slackness}.

\begin{itemize}
    \item \textbf{No Slackness:} Both the mug and the branch are explicitly specified. For example, \textit{``Put Bob's mug, the red mug, on the top branch.''} The task is only considered successful if the specified mug is placed on the designated branch.
    
    \item \textbf{Mug Slackness:} Only the branch is explicitly specified. For example, \textit{``Put away a mug on the right branch.''} Any mug can be selected, but it must be placed on the designated branch for the task to be considered successful.
    
    \item \textbf{Branch Slackness:} Only the mug is explicitly specified. For example, \textit{``Hang the green mug on a branch. Sorry, the blue mug.''} The specified mug must be picked up, but it can be placed on any available branch for the task to be considered successful.
    
    \item \textbf{Both Slackness:} Neither the mug nor the branch is explicitly specified. For example, \textit{``Hang a mug on a branch.''} In this case, placing any mug on any branch is considered successful.
\end{itemize}

\paragraph{Stowing a Book}  
A fixed number of books are placed on the bookshelf, with 1 to 2 of them tilted, creating 2 to 4 potential slots for stowing. Except for the 4 large books on the sides of the shelf, the order of the other books are randomized. There is also another randomly positioned book on a orange pad for picking. The goal of this task is to pick up the book on pad and stow it into a slot in the book shelf according to a given linguistic instruction.

We define three types of instructions based on the level of specificity: \textbf{no slackness, side slackness}, and \textbf{range slackness}.

\begin{itemize}
    \item \textbf{No Slackness:} The slot is explicitly specified. For example, \textit{``Place the front-most book so it sits directly next to DUBAI on the right side.''} The task is only considered successful if the book on the pad is stowed in the designated slot.
    
    \item \textbf{Side Slackness:} The book should be placed on the specified side of the specified book. For example, \textit{``Ensure LOST KINGDOMS is shelved on the orange book’s right side. Sorry, the left side.''}
    
    \item \textbf{Range Slackness:} The target slot is between two specified books. For example, \textit{``Put the Neo's book, Istanbul, onto the shelf so that it is between Urbanlike and the black book.''}
\end{itemize}

\subsection{Baseline Details}

\begin{itemize}
    \item \textbf{Lang-DP (PCD with color)}: The architecture of this baseline is essentially the same as Lang-DP (PCD) in the simulation experiments. The only difference is that the observation includes not only the point cloud of the scene but also additional RGB color channels for each point. In other words, each point consists of three channels representing its spatial location and three additional channels representing its color.
    \item \textbf{DP (PCD)}: It is similar to the Lang-DP (PCD) baseline in the simulation experiments. However, it can only observe raw point cloud without any hint from language.
    \item \textbf{Lang-DP (RGB)}: It is the same as the Lang-DP (RGB) used in the simulation experiments.
\end{itemize}

\section{Comparisons with VoxPoser}

VoxPoser relies on off-the-shelf motion planners to execute low-level skills~\cite{huang2023voxposer} and assumes simplified dynamics. For instance, once an object is grasped, it is treated as rigidly attached to the end-effector. As a result, VoxPoser is limited in handling scenarios involving complex dynamics or contact-rich interactions.

\begin{figure}[htbp!]
  \centering
  \includegraphics[width=0.95\linewidth]{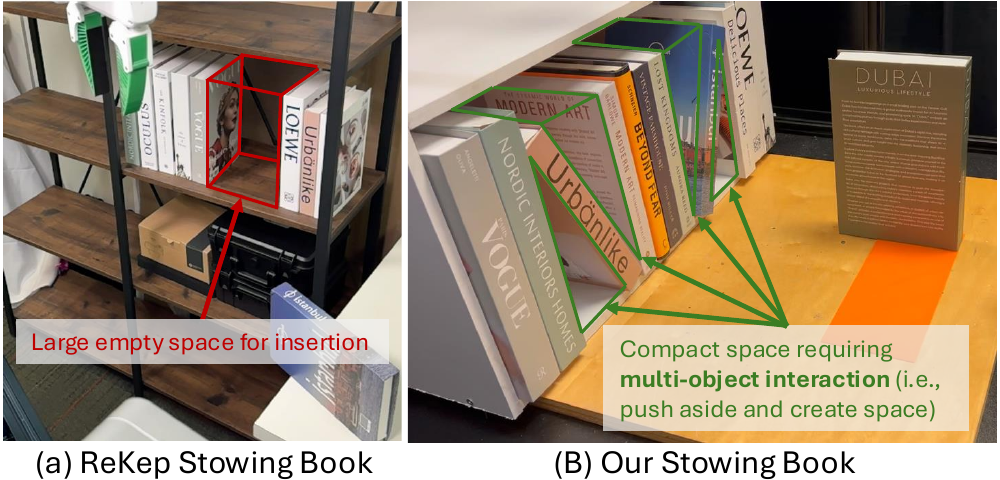}
  \caption{\small
  \textbf{Experiment Setup Comparison.}
  ReKep's setup assumes an empty slot for book insertion, whereas our setup requires complex multi-object interaction to push books aside and create space.
  }
  \label{fig:voxposer_comp}
\end{figure}

In contrast, many of the tasks in our work—such as book stowing—involve multi-object interactions and complex contact dynamics. As illustrated in Figure~\ref{fig:voxposer_comp}, prior work assumes an empty slot for book insertion. In our task setting, however, the robot must push adjacent books aside and squeeze out space—capabilities made possible by learning visuomotor policies from demonstrations.

\begin{figure}
  \centering
  \includegraphics[width=\linewidth]{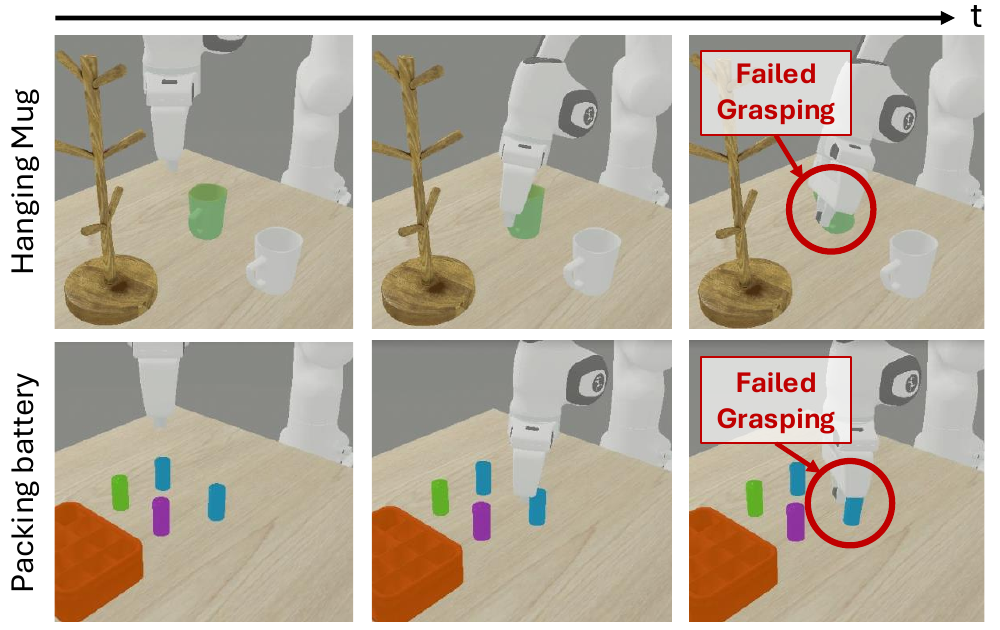}
  \caption{\small
  \textbf{VoxPoser Rollout Examples.} VoxPoser rollouts in our simulation environments demonstrate that VoxPoser fails to grasp the target objects.
  }
  \label{fig:voxposer}
\end{figure}

We also conduct additional experiments in simulation, with example rollouts shown in Figure~\ref{fig:voxposer}. Since VoxPoser relies on a pre-trained grasp planner, it may fail in scenarios that are not well-supported in the training data.

\bibliographystyle{unsrtnat}
\bibliography{reference}

\clearpage